\documentclass{article}

\usepackage{microtype}
\usepackage{graphicx}
\usepackage{subfigure}
\usepackage{booktabs} 

\usepackage{hyperref}

\usepackage[accepted]{icml2025}

\usepackage{amsmath}
\usepackage{amssymb}
\usepackage{mathtools}
\usepackage{amsthm}
\usepackage{multirow}
\usepackage{bbding}
\usepackage{makecell}
\usepackage[capitalize,noabbrev]{cleveref}

\theoremstyle{plain}

\theoremstyle{definition}

\theoremstyle{remark}

\usepackage[disable,textsize=tiny]{todonotes}
\usepackage[textsize=tiny]{todonotes}

\icmltitlerunning{Visual Autoregressive Modeling for Image Super-Resolution}

\begin{document}

\twocolumn[
\icmltitle{Visual Autoregressive Modeling for Image Super-Resolution}



\icmlsetsymbol{equal}{*}

\begin{icmlauthorlist}
\icmlauthor{Yunpeng Qu}{sch,comp}
\icmlauthor{Kun Yuan}{comp}
\icmlauthor{Jinhua Hao}{comp}
\icmlauthor{Kai Zhao}{comp}
\icmlauthor{Qizhi Xie}{sch,comp}
\icmlauthor{Ming Sun}{comp}
\icmlauthor{Chao Zhou}{comp}
\end{icmlauthorlist}

\icmlaffiliation{sch}{Tsinghua University, Beijing, China}
\icmlaffiliation{comp}{Kuaishou Technology, Beijing, China}

\icmlcorrespondingauthor{Kun Yuan}{yuankun03@kuaishou.com}

\icmlkeywords{Machine Learning, ICML}

\vskip 0.3in
]



\printAffiliationsAndNotice{} 

\begin{abstract}
Image Super-Resolution (ISR) has seen significant progress with the introduction of remarkable generative models.
However, challenges such as the trade-off issues between fidelity and realism, as well as computational complexity, have also posed limitations on their application.
Building upon the tremendous success of autoregressive models in the language domain, we propose \textbf{VARSR}, a novel visual autoregressive modeling for ISR framework with the form of next-scale prediction.
To effectively integrate and preserve semantic information in low-resolution images, we propose using prefix tokens to incorporate the condition.
Scale-aligned Rotary Positional Encodings are introduced to capture spatial structures and the diffusion refiner is utilized for modeling quantization residual loss to achieve pixel-level fidelity.
Image-based Classifier-free Guidance is proposed to guide the generation of more realistic images.
Furthermore, we collect large-scale data and design a training process to obtain robust generative priors.
Quantitative and qualitative results show that VARSR is capable of generating high-fidelity and high-realism images with more efficiency than diffusion-based methods.
Our codes will be released at \url{https://github.com/qyp2000/VARSR}.
\end{abstract}

\section{Introduction}





Image Super-Resolution (ISR) aims to generate realistic high-resolution (HR) images from their degraded low-resolution (LR) counterparts.
Traditional ISR methods focus on restoring LR images by assuming simple and known degradations  \cite{DBLP:conf/eccv/DongLT16,DBLP:conf/eccv/ZhangLLWZF18}, which limits their practicality in real-world scenarios with complex distortions \cite{DBLP:conf/cvpr/GuLZD19, DBLP:conf/icml/ZhangGCDZY23}.
Recent methods leverage generative model priors to tackle ISR, with approaches based on GAN models \cite{DBLP:conf/iccv/0008LGT21, DBLP:conf/iccvw/WangXDS21} and diffusion models \cite{DBLP:conf/nips/HoJA20}.
While these methods have achieved significant advancements, ISR, as an ill-posed problem, struggles to balance the realism and fidelity of the restoration results.
GAN-based methods achieve higher fidelity metrics, but limitations in generation capability and the goal of over-fidelity optimization make it challenging to reproduce vivid and realistic textures \cite{DBLP:journals/corr/abs-2311-16518}.
Diffusion methods, based on their strong generative priors, can generate extremely rich image details. 
However, the random noise sampling approach and the gap between generative priors and LR distribution pose challenges for pixel-level fidelity \cite{yu2024scaling}.
Building upon previous works, we aim to explore further existing frameworks to enhance fidelity and realism.

Introducing the generative capability of large models to enhance ISR is a prevalent trend.
With the rich semantic priors of T2I models pretrained on massive datasets, many works \cite{yang2025pixel, wu2024seesr} have applied the powerful generative models to ISR tasks and achieved significant results.
However, for ISR tasks demanding pixel-level fidelity, the diffusion model presents certain limitations, including the high computational complexity of iterative inference, and potential issues with semantic hallucinations \cite{kim2025tackling}.
With the success of autoregressive (AR) modeling in Large Language Models (LLMs) \cite{touvron2023llama, achiam2023gpt}, visual autoregressive modeling (VAR) has gained great attention, exemplified by VQGAN \cite{esser2021taming} and DALL-E \cite{reddy2021dall} which apply discrete token prediction for image generation.
VAR \cite{var} takes a step further by quantizing images into scale-wise token maps and generating images through next-scale prediction, leading to excellent results across multiple generation tasks \cite{tang2024hart, zhang2024var, yao2024car}.
This novel approach provides new insights into ISR, showcasing potential advantages over diffusion methods:
(1) The coarse-to-fine next-scale prediction conforms to Markov unidirectional modeling, providing better structural preservation and naturally adapting to ISR tasks; (2) The reduced number of inference steps and lower complexity in preceding scales result in higher efficiency.

However, unlike other controllable generation tasks, ISR requires generation with pixel-level fidelity alongside semantics reservation,
presenting challenges in integrating VAR into ISR tasks:
(1) How to efficiently and effectively incorporate LR condition information to generate high-fidelity images?
(2) AR-based approaches map images into 1D token sequences for processing. To achieve enhanced spatial structure preservation in ISR, how to better represent the positional relationships between tokens?
(3) AR-based approaches quantize continuous latent to discrete tokens for training with cross-entropy loss.
How to avoid the loss of information caused by the quantization process to improve the pixel-level fidelity of ISR? 
(4) Optimization objectives for fidelity may impact the generation of rich image details.
How to perceive low-quality distortions and guide the generation of more realistic images within the AR framework?


In this paper, we propose a novel \textbf{Visual Autoregressive Modeling for Image Super-Resolution (VARSR)} framework, with next-scale prediction.
To enhance the integration of semantic information from LR images, we propose using \textit{Prefix Tokens} as a proceeding scale to incorporate the LR condition as global guidance.
To better capture the spatial structure between tokens, we utilize the \textit{Scale-aligned Rotary Positional Encodings (SA-RoPE)} to calibrate the spatial positions of LR images and token maps.
To predict the residual loss of the discrete quantizer, the \textit{Diffusion Refiner} is introduced to model continuous distributions and achieve finer pixel-level restoration.
To generate more realistic images, we define negative samples to learn low-quality distortions and propose \textit{Image-based Classifier-free Guidance (CFG)} to guide the distribution for generating richer textures.
Furthermore, to leverage powerful generative priors, we collect large-scale data and plan a comprehensive training process.
Our \textbf{contributions} are as follows:
\begin{itemize}
    \item[1.] We first introduce VAR into the ISR field and propose VARSR, which is specifically designed to address LR conditions, spatial structure representation, quantization loss prediction, image-based CFG, and other issues to be applicable to ISR tasks.
    \item [2.] We collect a large-scale, high-quality image dataset to establish a robust base model and design a training pipeline for fine-tuning downstream ISR tasks. 
    \item[3.] Through quantitative and qualitative analysis, VARSR demonstrates a strong capability in generating high-fidelity and high-realism images, achieving \textbf{the best performance} on multiple image quality metrics and \textbf{$10\times$ efficiency improvement} of diffusion methods.
\end{itemize}

\section{Related work}
\paragraph{\textbf{\textit{Generative Priors for ISR.}}} 
Existing ISR methods typically focus on blind recovery without assuming specific degradations \cite{huang2020unfolding, bell2019blind}.
Generative priors are essential for addressing severe degradation in ISR tasks, capturing image structure and real-world distribution \cite{DBLP:conf/nips/HoJA20, DBLP:conf/cvpr/RombachBLEO22, DBLP:journals/corr/abs-2302-08453}.
While GAN-based methods show impressive results \cite{chen2022real,DBLP:conf/iccv/0008LGT21,DBLP:conf/iccvw/WangXDS21}, training stability and generative priors limit their ability to produce realistic details.
Diffusion models are popular in ISR for strong generative capabilities \cite{saharia2022image, kawar2022denoising, qu2025xpsr}, but may generate unrealistic details or hallucinations \cite{aithal2024understanding, narasimhaswamy2024handiffuser}.
This may stem from their simplification of the Markov process, restricting the access to antecedent denoised trajectories \cite{gu2024dart}.

\paragraph{\textbf{\textit{Visual Autoregressive Models.}}} 
Building upon the tremendous success of LLMs \cite{touvron2023llama, achiam2023gpt}, visual autoregressive models utilize discrete quantizers such as VQVAE \cite{van2017neural} to transform image patches into index-wise tokens, generating images based on next-token prediction \cite{yu2022scaling, wang2024emu3, lee2022autoregressive,chang2022maskgit}.
However, the prediction of flattened tokens may lead to the loss of spatial structure. 
VAR \cite{var} shifts from the next-token prediction to the next-scale prediction, significantly enhancing image generation quality and offering excellent scalability.
VAR-based methods such as STAR \cite{ma2024star}, ControlAR \cite{li2024controlvar} have expanded to other conditional generation tasks (\textit{e.g.}, class-to-image (C2I), text-to-image (T2I)), yielding results comparable to diffusion models and validating the potential of VAR \cite{zhang2024var, yao2024car, roheda2024cart, chen2024next}.

\section{Methods}
\subsection{Preliminary: visual autoregressive modeling}
\textbf{Next-token Prediction} is relied on by traditional AR models to generate patches at different positions in an image.
Images are mapped into latent by a visual autoencoder and then tokenized into a series of token maps $(x_1, x_2, ..., x_T)$ using a discrete quantizer.
AR model predicts the token $x_T$ based on its preceding sequence $(x_1, x_2, ..., x_{T-1})$ and condition $c$.
The conditional probability can be expressed:
\begin{equation}
\label{eq:token}
    p\left(x_1, x_2, ..., x_T\right)=\prod_{t=1}^T p\left(x_t \mid x_1, x_2, ..., x_{t-1}, c\right),
\end{equation}
Previous work \cite{var} has pointed out that “next-token prediction” is insufficient for highly structured images, as it will disrupt the spatial structure and defy the unidirectional dependence assumption of autoregression.

\textbf{Next-scale prediction} is reformulated based on these analyses \cite{var}, where the basic unit of autoregression becomes the token map of the entire scale. 
The visual encoder (\textit{e.g.}, VQVAE) first embeds the image $I$ into a feature map $f \in \mathcal{R}^{h\times w \times d}$ and then quantizes the feature map $f$ into $K$ multi-scale token maps $(r_1, r_2, ..., r_K)$.
\vspace{-0.5mm}
\begin{equation}
\label{eq:scale1}
    \begin{split}
    f_k = f - \sum_{m=1}^{k-1} upsample(\mathrm{lookup}(V, r_m)),\\
    r_k^{(i, j)}=\underset{v \in[V]}{\arg \min }\left\|\operatorname{lookup}(V, v)-f_k^{(i, j)}\right\|_2,\\
    \end{split}
    \vspace{-10mm}
\end{equation}

where $V$ is the VQVAE codebook, $r_k \in [V]^{h_k\times w_k}$ is the token map at scale $k$, and $\mathrm{lookup}(V, v)$ retrieves vectors from codebook $V$ based on index $v$.
VAR predicts the next scale $r_K$ based on previous outcomes and the condition:
\begin{equation}
\label{eq:scale2}
    p\left(r_1, r_2, ..., r_K\right)=\prod_{k=1}^K p\left(r_k \mid r_1, r_2, ..., r_{k-1}, c\right),
\end{equation}
VAR with next-scale prediction advances AR models greatly.
For ISR tasks demanding fidelity and realism, VAR's progressive generation with coarse-to-fine coherence aligns with human perception and the Markov unidirectional assumption, ensuring high structural fidelity and aesthetic quality.
Hence, applying VAR to ISR has broad prospects.


\subsection{VARSR framework}
\begin{figure*}[t]
  \centering
    \includegraphics[width=0.9\linewidth]{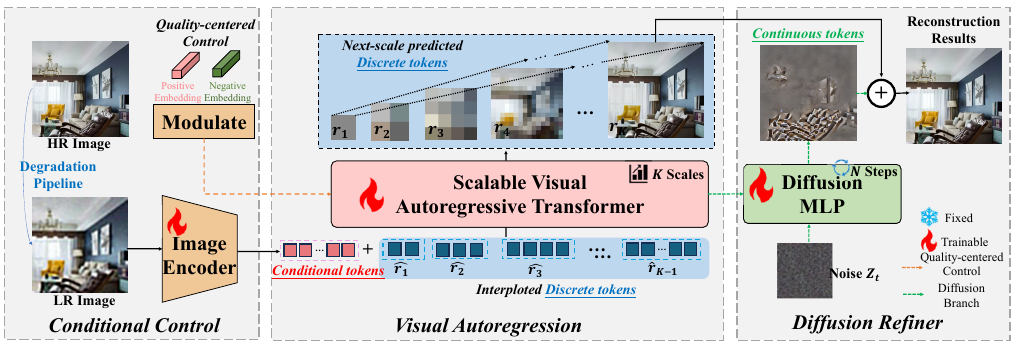}
    \vspace{-4mm}
  \caption{VARSR framework, which can be divided into three parts: 
  (1) LR image is set as \textit{Prefix Tokens} as condition. (2) VAR generates \textit{discrete tokens} based on next-scale prediction. (3) Diffusion Refiner predicts the \textit{continious tokens} as quantization residuals.}
  \label{fig:model}
  \vspace{-3mm}
\end{figure*}
We aim to enhance ISR via the VAR models.
Unlike other controllable generation tasks (\textit{e.g.}, C2I, T2I), ISR necessitates maintaining semantic integrity and achieving precise pixel-level restoration.
Therefore, when using the VAR, there still exist four critical issues requiring resolution:
\vspace{-1.5mm}
\begin{itemize}
    \item 
    In what manner can the conditions derived from LR images be efficiently and effectively furnished to VAR to achieve high-fidelity semantic restoration?
    \item 
    Tokens are concatenated into a 1D sequence in VAR, leading to the loss of spatial structures, which affects the fine-grained local restoration. How can the positional relationships of tokens be represented?
    \item
    VAR processes the image into discrete tokens through quantization, resulting in the significant loss of high-frequency details and impacting the pixel-level restoration. 
    How to reduce the impact of quantization loss?
    \item
    The fidelity-oriented optimization objective may affect the generation of image details.
    How can we perceive distortions and quality factors in images to generate more realistic and higher-quality images?
\end{itemize}
\vspace{-1.5mm}

To address these problems, we propose a framework called \textbf{Visual Autoregressive Modeling for Super Resolution (VARSR)}.
As depicted in Fig. \ref{fig:model}, it consists of three main stages: conditional control generation, visual autoregression, and diffusion refinement.
In the first stage, \textbf{\textcolor{red}{to address the 1st problem}}, we use an image encoder to map LR images to conditional tokens, which guide the generation of VAR by serving as \textit{Prefix Tokens} (Sec. \ref{sec:con}).
In the second stage, a pretrained VAR model is employed as the backbone.
\textbf{\textcolor{red}{To address the 2nd problem}}, we implement \textit{Scale-align rotary positional encoding (SA-RoPE)} in the transformer to calibrate the spatial positions of tokens at different scales (Sec. \ref{sec:block}).
In the third stage, \textbf{\textcolor{red}{to address the 3rd problem}}, a lightweight \textit{Diffusion Refiner} estimates quantization residuals from generated discrete tokens to enhance image details (Sec. \ref{sec:diff}).
\textbf{\textcolor{red}{To address the 4th problem}}, we further propose \textit{Image-based classifier-free guidance (CFG)} as an additional control to learn distortions and generate better-quality images without additional training (Sec. \ref{sec:cfg}).

\subsubsection{Conditional Control}
\label{sec:con}
The key challenge in ISR is how to effectively integrate LR information for controllable generation.
Diffusion methods often use ControlNet \cite{zhang2023adding} due to its superior performance over simpler approaches like concatenation or addition with noise.
However, for AR models, ControlNet increases the computational burden and may lead to conflicts between LR priors and prefix-scale information.

Considering the next-scale prediction form of VAR, we adopt a more efficient way to introduce LR conditional control called \textit{Prefix Tokens}. 
LR image $I_{lr}$ is encoded through an image encoder $\mathcal{E}$ to be mapped into conditional tokens $r_c=\mathcal{E}(I_{lr})$. 
$r_c$ is used as the initial scale and placed at the start of all tokens in the form $(r_c, r_1, r_2, ..., r_K)$, where $r_c$ is of size $h\times w$ like the final scale $r_K$.
VAR iteratively predicts the next scale by considering conditional tokens and the previous scales to effectively integrate LR priors.

In addition, HR images are categorized into two classes based on their quality during training. 
High- and low-quality categories are respectively associated with a positive embedding or a negative embedding for \textit{quality-centered control} in VAR through a modulate layer.
The objective is to impose \textit{Image-based CFG}, as will be elaborated on in Sec. \ref{sec:cfg}.

\subsubsection{Autoregressive Transformer}
\label{sec:block}
\begin{figure}[t]
  \centering
    \includegraphics[width=0.85\linewidth]{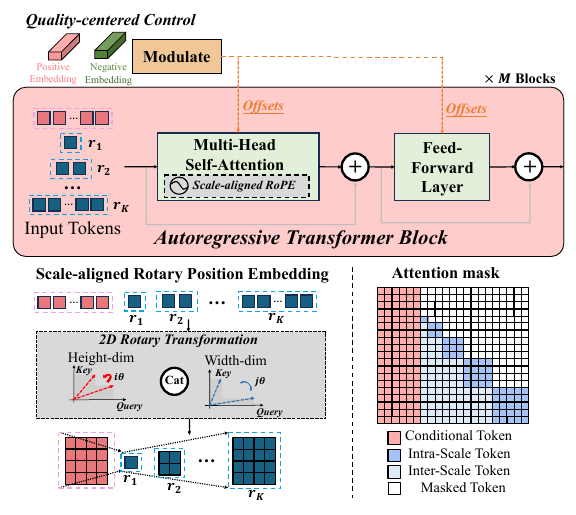}
      \vspace{-5mm}
  \caption{Internal structure of the autoregressive transformer. \textit{SA-RoPE} represents the spatial structure. \textit{Quality-centered control} generates offsets for autoregression.
}
  \label{fig:transformer}
  \vspace{-4mm}
\end{figure}
As shown in Fig. \ref{fig:transformer}, our VARSR utilizes a visual transformer for predicting multi-scale tokens.
In training, the tokens $r_{k}$ of the previous $K-1$ scales are interpolated to next scale:
\begin{equation}
\label{eq:scale3}
    \begin{split}
    \hat{r}_k = interpolate(\mathrm{lookup}(V, r_k)) \in \mathcal{R}^{h_{k+1}\times w_{k+1} \times d},\\
    \end{split}
\end{equation}
Conditional tokens and interpolated tokens are concatenated as $(r_c, \hat{r}_1, \hat{r}_2, ..., \hat{r}_{K-1})$ to predict the corresponding output $(\hat{r}_1, \hat{r}_2, ..., \hat{r}_K)$ for each scale.
A block-wise attention mask confines each $r_k$ to focus solely on its preceding tokens $r{\leq k}$, with $r_c$ accessible to all tokens to offer LR priors.


Transformers commonly use absolute positional encodings (APE) to represent the spatial relationships between tokens.
However, APE cannot capture 2D structure or represent the inter-scale positional relationships, which is crucial in ISR due to the challenge of incorporating spatial structure from LR images.
Inspired by previous works using RoPE \cite{su2024roformer} to represent the positions of token maps \cite{ma2024star, tang2024hart}, we propose an advancement with 2D RoPE to align spatial structures of LR images and multi-scale token maps, termed \textit{Scale-aligned RoPE}.
For the embedding $x_{k}^{(i,j)} \in \mathcal{R}^{C}$ of token $r_{k}^{(i,j)}$ in scale $k$ at position $(i,j)$, we split the channels into two for representing 2D positions and normalize encodings for scale alignment:
\begin{equation}
\label{eq:scale4}
    \begin{split}
    RoPE(x_{k}^{(i,j)}) = [ \begin{array}{cc}
    \mathbf{R}^{\frac{C}{2}}_{\Theta, (\frac{iH}{h_k})}     &  \mathbf{0}_{\frac{C}{2}}\\
    \mathbf{0}_{\frac{C}{2}}     &  \mathbf{R}^{\frac{C}{2}}_{\Theta, (\frac{jW}{w_k})}
    \end{array} ],
    \end{split}
\end{equation}
where $\mathbf{R}^C_{\Theta}, \mathbf{0}_{C} \in \mathcal{R}^{C \times C}$ are the standard rotation matrix for RoPE and the zero matrices, respectively.
Our \textit{SA-RoPE} will be applied to the multiplication in self-attention.
It is worth noting that we also perform scale alignment on the LR condition $r_{c}$ to incorporate the spatial structure of LR for next-scale restoration, enhancing the structural fidelity.

Lastly, the positive or negative embedding of the quality-centered control is introduced via the modulate layer, guiding the autoregression through the generation of offsets.

\subsubsection{Diffusion Refiner}
\label{sec:diff}
In Fig. \ref{fig:vqvae}, the image is quantized into discrete tokens, represented as indices of codebook $V$.
However, this discrete quantization introduces quantization loss, leading to the loss of high-frequency textures. 
Such loss is problematic for ISR which demands precise pixel-level recovery.
MAR \cite{li2024autoregressive} and HART \cite{tang2024hart} utilize a \text{diffusion loss} to map predicted tokens to a continuous probability distribution, mitigating quantization loss. 
VARSR adopts the above approach of using diffusion as a refiner.
The quantization residual is defined as continuous tokens $z$.
\begin{equation}
\label{eq:scale5}
    \begin{split}
    z = f - \sum_{k=1}^{K} upsample(\mathrm{lookup}(V, r_k)),\\
    \end{split}
\end{equation}
We introduce a simple \textit{Diffusion Refiner}, composed of a lightweight MLP \cite{li2024autoregressive}, to model continuous tokens in addition to predicting discrete tokens.
It leverages the final scale hidden states $x_K$ as conditional control to generate a continuous distribution from noise $z_t$.
\begin{equation}
\label{eq:scale6}
    \begin{split}
    \mathcal{L}(x_K, z)=\mathbb{E}_{\varepsilon \sim \mathcal{N}(0,1), t}\left[\left\|\varepsilon-\varepsilon_\theta\left(z_t \mid t, x_K\right)\right\|^2\right] ,\\
    \end{split}
\end{equation}
where $\varepsilon$ is the sampled noise, $t$ is the timestamp and $\varepsilon_\theta(\cdot)$ is the mapping function of the \textit{Diffusion Refiner}.
During inference, the sampled continuous tokens $\hat{z}$ are combined with the predicted token maps for the final output.

\subsubsection{Image-based Classifier-Free Guidance}
\label{sec:cfg}
Classifier-Free Guidance (CFG) \cite{ho2022classifier} has been widely used in T2I diffusion models, leveraging negative prompts to generate more realistic images by directed guiding of probabilistic distributions.
Instead of simply setting the condition to null \cite{li2024controlar, ho2022classifier} or using textual prompts, we aim to better perceive distortions and low-quality factors in images, hence proposing \textit{Image-based CFG} to enhance generation quality.

In training, HR images are divided into high- and low-quality classes, with positive and negative embeddings $c_{p}, c_{n}$ for control.
In inference, to generate a higher-quality image $I$ with condition $r_c$ and $c_{p}$, the distribution is:

\vspace{-5mm}
\begin{equation}
\label{eq:cfg0}
    \begin{split}
    &p(I|r_c,c_{p}) = \frac{p(c_{p}|r_c,I)p(I|r_{c})}{p(c_{p}| r_{c})},\\
    p(I|r_{c}&) = p(c_{p}|r_{c})p(I|r_c, c_{p}) + p(c_{n}|r_{c})p(I|r_c, c_{n}),
    \end{split}
\end{equation}
Eq. \ref{eq:cfg0} can be simplified to:
\begin{equation}
\label{eq:cfg1}
    \begin{split}
    p(I|r_c,c_{p}) = \frac{p(c_n|r_{c})}{p(c_p|r_{c})}\frac{p(c_p|r_c, I)p(I|r_c, c_n)}{1-p(c_p|r_c, I)},\\
    \Rightarrow \nabla_I \log p(I|r_c,c_{p})=\nabla_I \log \frac{p(c_p|r_c, I)}{1-p(c_p|r_c, I)} \\
    +\nabla_I \log p(I|r_c, c_n),
    \end{split}
\end{equation}
Furthermore, by applying  Bayesian formula, we can obtain:
\begin{equation}
\label{eq:cfg2}
    \begin{split}
    \frac{p(c_p|r_c, I)}{1-p(c_p|r_c, I)} = \frac{p(c_p|r_{c})}{p(c_n|r_{c})}\frac{p(I|r_c,c_p)}{p(I|r_c,c_n)},
    \end{split}
\end{equation}
Substituting Eq. \ref{eq:cfg2} into Eq. \ref{eq:cfg1} yields Eq. \ref{eq:cfg3}. In CFG, a guiding scale $\lambda_s$ balances diversity and realism.
\begin{equation}
\label{eq:cfg3}
    \begin{split}
    \nabla_I &\log p(I|r_c,c_{p}) = \nabla_I \log p(I|r_c, c_n)\\
   & +\lambda_s(\nabla_I \log p(I|r_c, c_p) -\nabla_I \log p(I|r_c, c_n)),
    \end{split}
\end{equation}
During inference, \textit{Image-based CFG} are represented as:
\begin{equation}
\label{eq:cfg4}
    \begin{split}
    &\widetilde{I} = \mathcal{F}(r_c, c_p) + \lambda_s(\mathcal{F}(r_c, c_p) -\mathcal{F}(r_c, c_n)).
    \end{split}
\end{equation}
where $\mathcal{F}(\cdot)$ is the mapping function of our VARSR.

\subsection{Scaling up Database}
\label{sec:data}
\paragraph{\textbf{\textit{Large-scale Data Collection.}}} 
ISR methods require ample data to acquire generative priors and semantic understanding capabilities\cite{yu2024scaling}. 
Diffusion methods use Stable Diffusion (SD) \cite{rombach2022high} trained on billions of text-image pairs for plentiful image priors. 
VAR lacks such powerful models, necessitating scaling high-quality training data for robust priors.
Classical datasets often struggle to meet our high standards for image quality (\textit{e.g.}, ImageNet \cite{deng2009imagenet}) or quantity (\textit{e.g.}, DIV2K \cite{DBLP:conf/cvpr/AgustssonT17}, DIV8K \cite{DBLP:conf/iccvw/GuLDFLT19}).
Therefore, we collect a \textbf{new large-scale dataset with 4 million high-quality, high-resolution images across over 3k categories}, ensuring rich details and clear semantics.

\vspace{-1mm}
\paragraph{\textbf{\textit{Negative Samples.}}} 
In Sec. \ref{sec:cfg}, our VARSR utilizes negative embedding as inverse control to generate low-quality images.
Therefore, we provide low-quality images as negative samples corresponding to negative embedding to learn low-quality distortions. 
We sample 50k low-quality images from various manually annotated image quality assessment (IQA) datasets (\textit{e.g.}, KonIQ10K \cite{kon10k}, CLIVE \cite{clive}) and image aesthetics assessment (IAA) dataset AVA \cite{ava} as negative samples added to our database.

\vspace{-1mm}
\paragraph{\textbf{\textit{Image Preprocess.}}} 
Psychophysics research suggests that the richness of details in visual content impacts human perception of quality and aesthetics \cite{hvs2}.
Therefore, instead of the commonly-used random cropping, we resize images to 1.25 times the size for input and center-crop them during training.
The aim is to enhance the comprehensive coverage of the foreground, which typically contains richer semantics and textures than the background, enabling VARSR to capture a broader range of visual attributes.


\subsection{Training Procedure}
\label{sec:train}
\begin{figure}[t]
  \centering
    \includegraphics[width=0.9\linewidth]{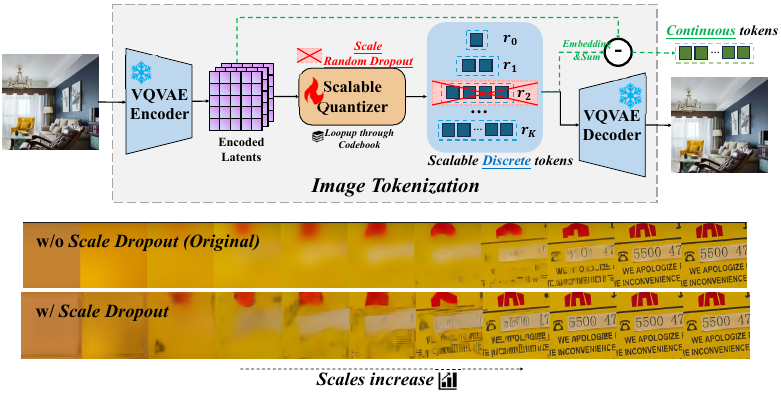}
    \vspace{-0.5cm}
  \caption{Image tokenization process of VQVAE. The quantizer converts the image latent to multi-scale \textit{discrete tokens} while representing the quantization loss as \textit{continuous tokens}.
}
  \label{fig:vqvae}
  \vspace{-0.5cm}
\end{figure}
\begin{table*}[t]
  \centering
  \scriptsize
  \caption{Comparison with SOTA methods on synthetic and real-world benchmarks. 
  \textcolor{red}{Red} and \textcolor{blue}{blue} colors are the best and second-best.
  }
  \label{tab:headings}
  \centering
  \begin{tabular}{c|c|ccc|ccccc|c}
    \toprule
    \multirow{2}{*}{Dataset} & \multirow{2}{*}{Metrics} & \multicolumn{3}{c|}{GAN-based} & \multicolumn{5}{c|}{Diffusion-based}& AR-based \\
    & & BSRGAN & Real-ESR & SwinIR & LDM & StableSR & DiffBIR & PASD & SeeSR & \textbf{VARSR} \\
    \midrule
    \multirow{9}{*}{\textit{DIV2K-Val}}
    & PSNR$\uparrow$     & \textcolor{red}{24.42}& \textcolor{blue}{24.30}& 23.77& 21.66& 23.26& 23.49& 23.59& 23.56& 23.91 \\
    & SSIM$\uparrow$     & 0.6164& \textcolor{red}{0.6324}& \textcolor{blue}{0.6186}& 0.4752& 0.5670& 0.5568& 0.5899& 0.5981&0.5980\\
    & LPIPS$\downarrow$  & 0.3511&0.3267& 0.3910& 0.4887& \textcolor{red}{0.3228}& 0.3638& 0.3611& 0.3283& \textcolor{blue}{0.3260} \\
    & DISTS$\downarrow$  & 0.2369& 0.2245& 0.2291& 0.2693& \textcolor{blue}{0.2116}& 0.2177& 0.2134&\textcolor{red}{0.2008}&0.2218 \\
    & FID$\downarrow$    & 50.99& 44.34& 44.45& 55.04& \textcolor{red}{28.32}& 34.55& 39.74& \textcolor{blue}{28.89}&35.51 \\
    & MANIQA$\uparrow$   & 0.3547& 0.3756& 0.3411& 0.3589& 0.4173& 0.4598& 0.4440& \textcolor{blue}{0.5046}&\textcolor{red}{0.5340} \\
    & CLIPIQA$\uparrow$  & 0.5253& 0.5205& 0.5213& 0.5570& 0.6752& 0.6731&0.6573& \textcolor{blue}{0.6959}&\textcolor{red}{0.7347} \\
    & MUSIQ$\uparrow$    & 60.18& 59.76& 57.21& 57.46& 65.19& 65.57& 66.58& \textcolor{blue}{68.35}&\textcolor{red}{71.27} \\

    \midrule
    \multirow{9}{*}{\textit{RealSR}}
    & PSNR$\uparrow$     & \textcolor{red}{26.38}& 25.68& \textcolor{blue}{25.88}& 25.66& 24.69& 24.94& 25.21& 25.31& 24.61 \\
    & SSIM$\uparrow$     & \textcolor{blue}{0.7651}& 0.7614& \textcolor{red}{0.7671}& 0.6934& 0.7090& 0.6664& 0.7140& 0.7284& 0.7169 \\
    & LPIPS$\downarrow$  & \textcolor{blue}{0.2656}& 0.2710& \textcolor{red}{0.2614}& 0.3367& 0.3003& 0.3485& 0.2986& 0.2993& 0.3504 \\
    & DISTS$\downarrow$  & 0.2124& \textcolor{red}{0.2060}& \textcolor{blue}{0.2061}& 0.2324& 0.2134& 0.2257& 0.2125& 0.2224& 0.2470 \\
    & FID$\downarrow$    & 141.25& 135.14& 132.80& 133.34& 131.72& \textcolor{blue}{127.59}& 139.42& \textcolor{red}{126.21}& 137.55 \\
    & MANIQA$\uparrow$   & 0.3763& 0.3736& 0.3561& 0.3375& 0.4167& 0.4378& 0.4418& \textcolor{blue}{0.5370}& \textcolor{red}{0.5570} \\
    & CLIPIQA$\uparrow$  & 0.5114& 0.4487& 0.4433& 0.6053& 0.6200& 0.6396& 0.6009& \textcolor{blue}{0.6638}& \textcolor{red}{0.7006} \\
    & MUSIQ$\uparrow$    & 63.28& 60.37& 59.28& 56.32& 65.25& 64.32& 66.61&\textcolor{blue}{ 69.56}& \textcolor{red}{71.26} \\
   
    \midrule
    \multirow{9}{*}{\textit{DRealSR}}
    & PSNR$\uparrow$     & \textcolor{red}{28.70} & \textcolor{blue}{28.61}& 28.20& 27.78& 27.87& 26.57& 27.45& 28.13& 28.16 \\
    & SSIM$\uparrow$     & \textcolor{blue}{0.8028}& \textcolor{red}{0.8052}& 0.7983& 0.7152&  0.7427& 0.6516& 0.7539& 0.7711& 0.7652 \\
    & LPIPS$\downarrow$  & 0.2858& \textcolor{red}{0.2819}& \textcolor{blue}{0.2830}& 0.3745& 0.3333& 0.4537& 0.3331& 0.3142&0.3541 \\
    & DISTS$\downarrow$  & 0.2144& \textcolor{red}{0.2089}& \textcolor{blue}{0.2103}& 0.2417& 0.2297& 0.2724& 0.2322& 0.2230&0.2526 \\
    & FID$\downarrow$    & 155.62& 147.66& \textcolor{red}{146.38}& 164.87& 148.18& 160.67& 173.40& \textcolor{blue}{147.00}&155.87 \\
    & MANIQA$\uparrow$   & 0.3441& 0.3435& 0.3311& 0.3342& 0.3897&0.4602& 0.4551& \textcolor{blue}{0.5077}& \textcolor{red}{0.5362} \\
    & CLIPIQA$\uparrow$  & 0.5061& 0.4525& 0.4522& 0.5984& 0.6321& 0.6445& 0.6365&\textcolor{blue}{0.6893} & \textcolor{red}{0.7240} \\
    & MUSIQ$\uparrow$    & 57.16& 54.27& 53.01& 51.37& 58.72&61.06& 63.69& \textcolor{blue}{64.75} & \textcolor{red}{68.15} \\
  \bottomrule
  \end{tabular}
  \vspace{-5mm}
\end{table*}


\paragraph{\textbf{\textit{VQVAE.}}} 
In Fig. \ref{fig:vqvae}, the visualizations of the original VAR quantizer \cite{var} reveal that image semantics are concentrated in the final few scales.
This limits ISR as earlier scale generation may lack information for later scales.
Hence, we follow the \textit{scale random dropout} strategy in previous works \cite{li2024imagefolder, kumar2024high}, where multi-scale quantized results are randomly discarded with a probability $p_d$.
We freeze other parts and only train the quantizer. 
In Fig. \ref{fig:vqvae}, applying scale dropout preserves more semantic information in the earlier scales.


\vspace{-1mm}
\paragraph{\textbf{\textit{C2I Pretraining.}}} 
We first train a powerful base VAR model on the C2I task to establish robust generation priors.
Our training utilizes our large-scale dataset of over 3k categories,  incorporating class information via start tokens and modulate layers in the transformer \cite{var}.

\vspace{-1mm}
\paragraph{\textbf{\textit{ISR Finetuning.}}} 
We fine-tune the C2I pre-trained base model for the downstream ISR task to create the final VARSR.
Both C2I Pretraining and ISR Finetuning stages utilize the same loss with a coefficient $\lambda$ to balance the cross-entropy loss of tokens and diffusion loss of the refiner.


\section{Experiments}
\subsection{Experimental Setups}
\paragraph{\textbf{\textit{Datasets.}}} 
We train VARSR on our large-scale dataset with negative samples using Real-ESRGAN's degradation pipeline \cite{DBLP:conf/iccvw/WangXDS21} to synthesize LR-HR image pairs.
Both synthetic and real-world datasets are utilized for a comprehensive evaluation.
We create the synthetic validation set \textit{DIV2K-VAL} by randomly cropping 3k patches from the DIV2K \cite{DBLP:conf/cvpr/AgustssonT17} validation set, and for real-world evaluation, \textit{DrealSR}  \cite{DBLP:conf/iccv/CaiZYC019} and \textit{RealSR} \cite{DBLP:conf/eccv/WeiXLZYZL20} are center-cropped.
Following \cite{DBLP:journals/corr/abs-2305-07015}, all HR images in training and testing have a resolution of $512\times512$ and LR images are $128\times128$.

\paragraph{\textbf{\textit{Implemental Details.}}} 
We train our VARSR following the procedure in Sec. \ref{sec:train}, using a GPT-2 style \cite{radford2019language} transformer with 24 blocks as the base model and a Diffusion Refiner with 6 blocks. 
We accelerate training by using pretrained VAR \cite{var}.
We utilize an AdamW \cite{loshchilov2017fixing} optimizer with batchsize=128, weight decay=5e-2, and learning rate=5e-5.
VQVAE, C2I pretraining, and ISR finetuning run for 10k, 40k, and 20k iterations respectively.
The loss balancing coefficient $\lambda$ is 2.0 and the dropout ratio $p_{d}$ is 0.1.
The guidance scale $\lambda_s$ linearly increases to $6.0$ as the scale increases.
Experiments are performed on 32 NVIDIA V100 GPUs.

\begin{figure}[t]
  \centering
    \includegraphics[width=0.85\linewidth]{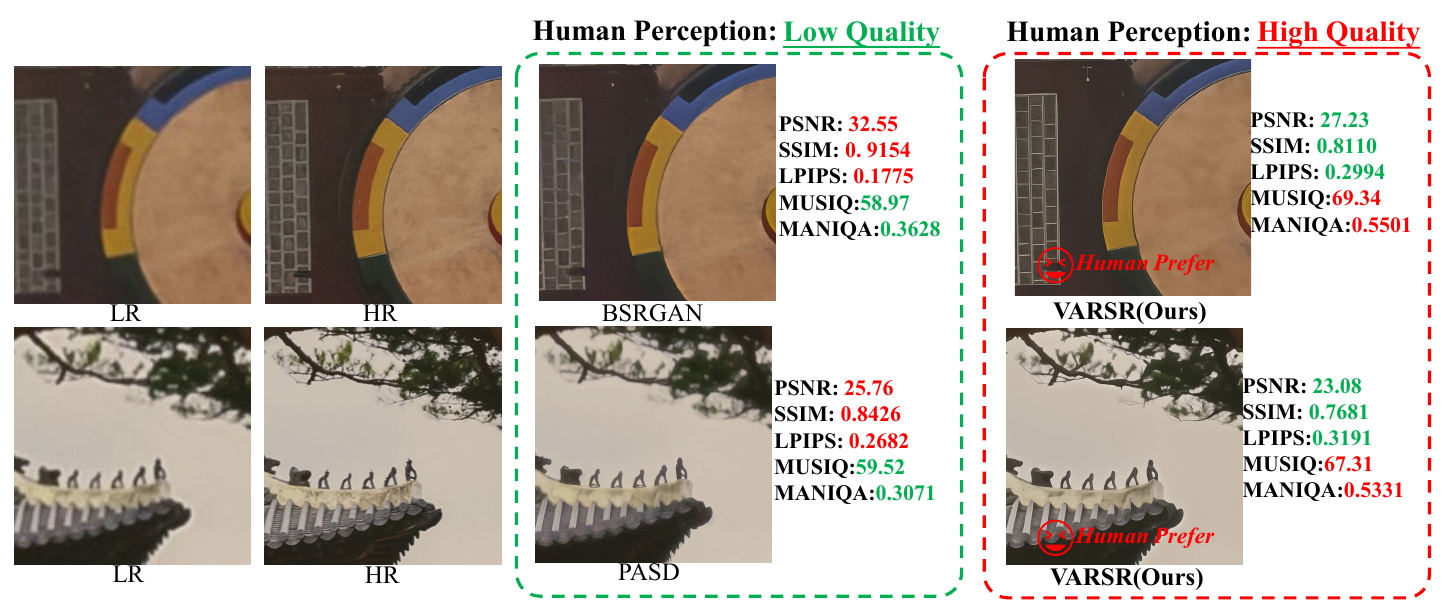}
    \vspace{-3mm}
  \caption{Limitations of current full-reference metrics (\textit{e.g.}, PSNR, SSIM, LPIPS). VARSR has generated images of higher perceptual quality for humans, yet it lags behind in certain metrics.}
  \label{fig:metrics}
  \vspace{-5mm}
\end{figure}

\paragraph{\textbf{\textit{Metrics.}}} 
Reference-based metrics including PSNR, SSIM  \cite{DBLP:journals/tip/WangBSS04} (Y channel),  \cite{DBLP:conf/cvpr/ZhangIESW18}, and DISTS \cite{DBLP:journals/pami/DingMWS22} are used for fidelity evaluation.
FID \cite{DBLP:conf/nips/HeuselRUNH17} measures the distribution distance between generated and reference images
MANIQA \cite{DBLP:conf/cvpr/YangWSLGCWY22}, MUSIQ \cite{ke2021musiq}, and CLIPIQA \cite{wang2023exploring} are non-reference IQA metrics.


\subsection{Comparison with SOTA}
\begin{figure*}[t]
  \centering
    \includegraphics[width=0.88\linewidth]{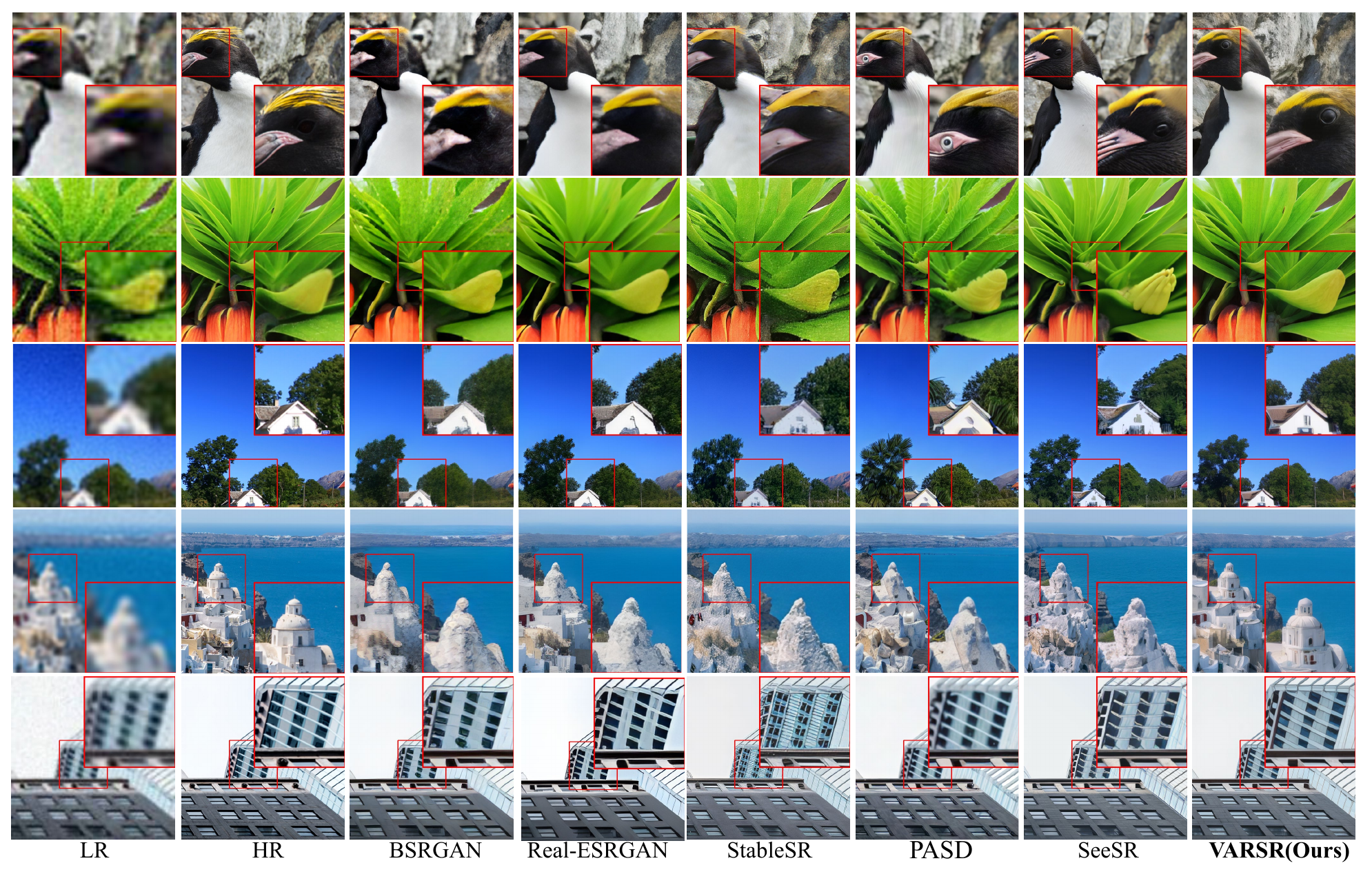}
    \vspace{-5mm}
    \caption{Qualitative comparisons with different SOTA methods. \textbf{Zoom in for a better view}.}
  \vspace{-5mm}
  \label{fig:sota}
\end{figure*}

\begin{table*}[t]
  \centering
  \caption{Ablation on the \textit{LR Condition Mode}.}
  \label{tab:lrcondition}
  \scriptsize
  \begin{tabular}{c|ccccc|ccccc}
    \toprule
    \multirow{2}{*}{\makecell{Condition \\ Mode}} & \multicolumn{5}{c|}{\textit{DrealSR}} & \multicolumn{5}{c}{\textit{RealSR}} \\
    ~ & SSIM$\uparrow$ &LPIPS$\downarrow$& DISTS$\downarrow$ & MANIQA$\uparrow$ & MUSIQ$\uparrow$
    & SSIM$\uparrow$  &LPIPS$\downarrow$&DISTS$\downarrow$ & MANIQA$\uparrow$ & MUSIQ$\uparrow$ \\
    \midrule
    Directly Add&  0.6902&0.4011&0.3010&0.5188&66.46
    & 0.6712 & 0.3742 & 0.2553 & 0.5221 & 69.90\\
    ControlNet& 0.7188 & 0.3824& 0.2752&  0.5260 & 66.78 & 0.7034 & 0.3683 & 0.2657 & 0.5298 & 69.70\\
    
    ControlNet+CA& 0.7314& 0.3784&0.2630& 0.5224 & 65.97 & 0.7084 & 0.3577& 0.2552 & 0.5270 & 69.83 \\
    Prefix Tokens& \textbf{0.7652} & \textbf{0.3541}&\textbf{0.2526}&\textbf{0.5362} &\textbf{68.15} & \textbf{0.7169}&\textbf{0.3504}&\textbf{0.2470}& \textbf{0.5570}& \textbf{71.26}\\
  \bottomrule
  \end{tabular}
  \vspace{-4mm}
\end{table*}

\begin{table}[t]
  \centering
  \caption{Complexity comparison.}
  \label{tab:complexity}
  \scriptsize
  \begin{tabular}{c|c|c|c}
    \toprule
Methods & Params & Steps & Inference Time\\
\midrule
StableSR & 1409.1M & 200 & 18.70s\\
PASD  & 1900.4M  &20  &6.07s\\
DiffBIR  & 1716.7M  & 50  &5.85s\\
SeeSR  & 2283.7M  &50 & 7.24s\\
\midrule
\textbf{VARSR(Ours)}  & \textbf{1101.9M}  & \textbf{10}  & \textbf{0.59s}\\
  \bottomrule
  \end{tabular}
  \vspace{-3mm}
\end{table}
\begin{figure}[t]
  \centering
    \includegraphics[width=0.8\linewidth]{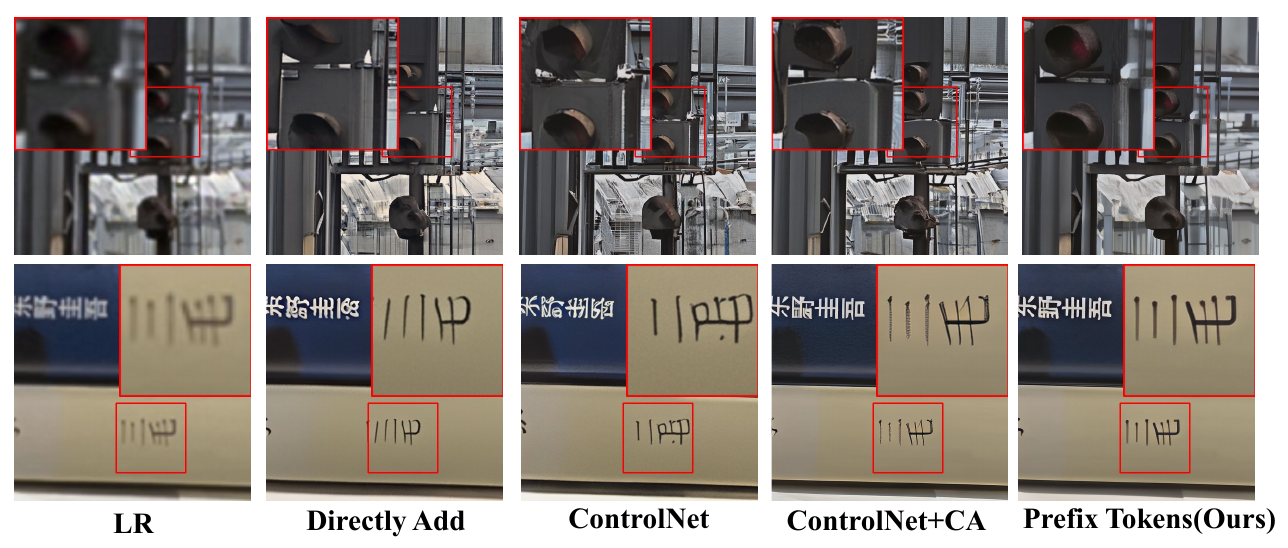}
  \vspace{-4mm}
  \caption{Effectivess of the \textit{LR Condition Mode}.}
  \vspace{-6mm}
  \label{fig:lrcondition}
\end{figure}

We compare our VARSR with other SOTA GAN-based and Diffusion-based ISR methods\footnote{All methods are tested based on their official code and models.}, including BSRGAN~\cite{DBLP:conf/iccv/0008LGT21}, Real-ESRGAN~\cite{DBLP:conf/iccvw/WangXDS21}, SwinIR-GAN~\cite{DBLP:conf/iccvw/LiangCSZGT21}, LDM~\cite{DBLP:conf/cvpr/RombachBLEO22}, StableSR~\cite{DBLP:journals/corr/abs-2305-07015}, DiffBIR~\cite{DBLP:journals/corr/abs-2308-15070}, PASD~\cite{yang2025pixel} and SeeSR \cite{wu2024seesr}.

\vspace{-1mm}
\paragraph{\textbf{\textit{Quantitative comparisons.}}} 
The quantitative comparisons are shown in Tab. \ref{tab:headings}.
\textcolor{red}{\textbf{Firstly}}, \textbf{VARSR outperforms other SOTA methods by a wide margin in three no-reference IQA metrics}: MAINIQA, CLIPIQA, and MUSIQ, reflecting the capability to produce high-quality and realistic images.
\textcolor{red}{\textbf{Secondly}}, in reference-based metrics (\textit{e.g.}, PSNR, SSIM), our VARSR approximates that of diffusion methods, yet still lags behind GAN methods.
This is due to the trade-off between realism and fidelity, as VARSR and diffusion methods generate more textures and details, which may reduce fidelity metrics, especially for lower-quality HR images.
In Fig. \ref{fig:metrics}, VARSR's restoration exhibits higher quality in human perception, yet lags behind in certain reference-based metrics.
This phenomenon highlights the limitations of current reference-based metrics, a concern confirmed by many previous works\cite{DBLP:journals/corr/abs-2305-07015,DBLP:journals/corr/abs-2311-16518}.

\paragraph{\textbf{\textit{Qualitative comparisons.}}} 

In Fig. \ref{fig:sota}, we visualize some of the ISR results from the test set.
\textcolor{red}{\textbf{Firstly}}, our VARSR consistently generates images with rich details and clear semantic information, showcasing the strong generative prior capability of autoregressive methods.
\textcolor{red}{\textbf{Secondly}}, the image quality generated by GAN-based methods lags behind diffusion-based methods and our AR-based VARSR, highlighting the significance of generative priors.
\textcolor{red}{\textbf{Thirdly}}, \textbf{VARSR also outperforms diffusion-based methods in detail generation capability, aligning with the IQA metrics in Tab. \ref{tab:headings}}.
For example, in the 2nd and 4th row, VARSR distinguishes itself as \textbf{the sole method} that can correctly understand the semantics and generate houses and leaves with clear textures.
In the 1st, 3rd, and 5th rows, VARSR demonstrates superior structural fidelity over other methods, generating images with well-defined structures and clear semantics.
Additionally, VARSR excels in creating intricate animal fur, detailed leaves, and vivid textures.
The results highlight VARSR's robust ability to produce realistic and semantic-preserved images, even under severe degradation of LR images.

\paragraph{\textbf{\textit{Complexity analysis.}}}
Diffusion models have longer inference times due to multi-step noise sampling, while VARSR generates token maps at different scales with low iteration counts, ensuring minimal inference time.
Additionally, with fewer token counts at early scales, the complexity of earlier steps is very low,  unlike diffusion methods with consistent complexity per iteration.
In Tab. \ref{tab:complexity}, compared to diffusion models, \textbf{VARSR only requires 0.59s to generate an image, which is 10.1\% of the second-ranked DiffBIR}.

\begin{table}[t]
    \begin{minipage}[t]{\linewidth}
		\centering
        \scriptsize
        \vspace{-2mm}
        \caption{Ablation on the \textit{Scale-aligned RoPE}.}
      \label{tab:rope}
       \centering
       \begin{tabular}{c|ccc|ccc}
        \toprule
        Exp &\multicolumn{3}{c|}{\textit{DrealSR}} & \multicolumn{3}{c}{\textit{RealSR}} \\
        \midrule
         Condition&  \XSolidBrush &  \XSolidBrush &\checkmark& \XSolidBrush &  \XSolidBrush &\checkmark\\
        Discrete&  \XSolidBrush &  \checkmark &\checkmark& \XSolidBrush &  \XSolidBrush &\checkmark\\
        \midrule
        SSIM$\uparrow$ &  0.7424& 0.7603& \textbf{0.7652} & 0.6992& 0.7039 & \textbf{0.7169}\\
        LPIPS$\downarrow$ & 0.3691& 0.3645& \textbf{0.3541} & 0.3654& 0.3672&\textbf{0.3504} \\
        DISTS$\downarrow$ &0.2617 &0.2642 &\textbf{0.2526} &0.2552 & 0.2576 &\textbf{0.2470} \\
        MANIQA$\uparrow$ & 0.5238 & \textbf{0.5490} & 0.5362  &  0.5307 & 0.5555&\textbf{0.5570} \\
        MUSIQ$\uparrow$ & 68.14& \textbf{68.41} &68.15 & 71.08 & \textbf{71.32}& 71.26 \\
      \bottomrule
      \end{tabular}
    \end{minipage}\\
    \begin{minipage}[t]{\linewidth}
		\centering
        \scriptsize
        \vspace{-1mm}
        \caption{Ablation on the \textit{Diffusion Refiner}.}
		\label{tab:diff}
        \begin{tabular}{c|cc|cc}
    \toprule
    \multirow{2}{*}{Metrics} &\multicolumn{2}{c|}{\textit{DrealSR}} & \multicolumn{2}{c}{\textit{RealSR}} \\
    & w/o & Diff. & w/o &  Diff.\\
    \midrule
    SSIM$\uparrow$ &  0.7583 & \textbf{0.7652} & 0.7118 & \textbf{0.7169}\\
    LPIPS$\downarrow$ & \textbf{0.3532} & 0.3541 & 0.3515 &\textbf{0.3504} \\
    DISTS$\downarrow$ &0.2539 &\textbf{0.2526} & 0.2503 &\textbf{0.2470} \\
    MANIQA$\uparrow$ & 0.5297 & \textbf{0.5362}  &  0.5399 &\textbf{0.5570} \\
    MUSIQ$\uparrow$ & 67.99 &\textbf{68.15} & 70.74 & \textbf{71.26} \\
  \bottomrule
  \end{tabular}
    \end{minipage}\\
    \begin{minipage}[t]{\linewidth}
          \centering
            \scriptsize
            \vspace{-1mm}
          \caption{Ablation on the \textit{Image-based CFG}.}
          \label{tab:cfg}
           \centering
          \begin{tabular}{c|cc|cc}
            \toprule
            \multirow{2}{*}{Metrics} &\multicolumn{2}{c|}{\textit{DrealSR}} & \multicolumn{2}{c}{\textit{RealSR}} \\
            & w/o & w/ CFG & w/o & w/ CFG\\
            \midrule
            SSIM$\uparrow$ & \textbf{0.8004} & 0.7652 & \textbf{0.7436} & 0.7169\\
            LPIPS$\downarrow$ &\textbf{0.2961} &0.3541 & \textbf{0.2977} &0.3504 \\
            DISTS$\downarrow$ &\textbf{0.2161} &0.2526 & \textbf{0.2103} &0.2470 \\
            MANIQA$\uparrow$ &  0.4052& \textbf{0.5362}  & 0.4326 &\textbf{0.5570} \\
            MUSIQ$\uparrow$ & 59.70 &\textbf{67.63} & 66.32 & \textbf{71.26} \\
          \bottomrule
          \end{tabular}
    \end{minipage}
    \vspace{-6mm}
\end{table}

\begin{figure}[t]
    \begin{minipage}[t]{\linewidth}
		  \centering
        \includegraphics[width=0.81\linewidth]{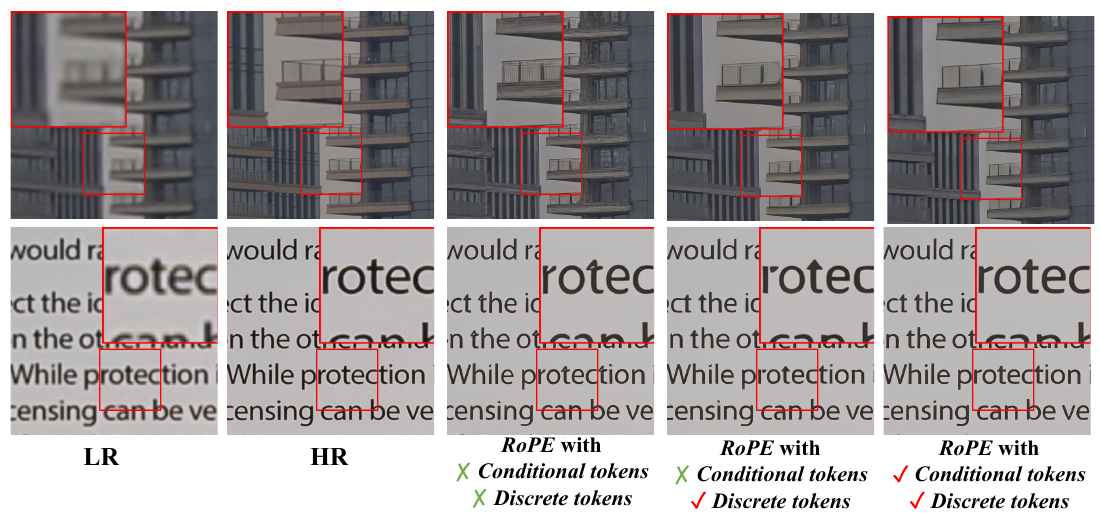}
      \vspace{-4.5mm}
      \caption{Effectivess of the \textit{Scale-aligned RoPE}.}
      \label{fig:rope}
    \end{minipage}\\
    \begin{minipage}[t]{\linewidth}
		  \centering
        \includegraphics[width=0.81\linewidth]{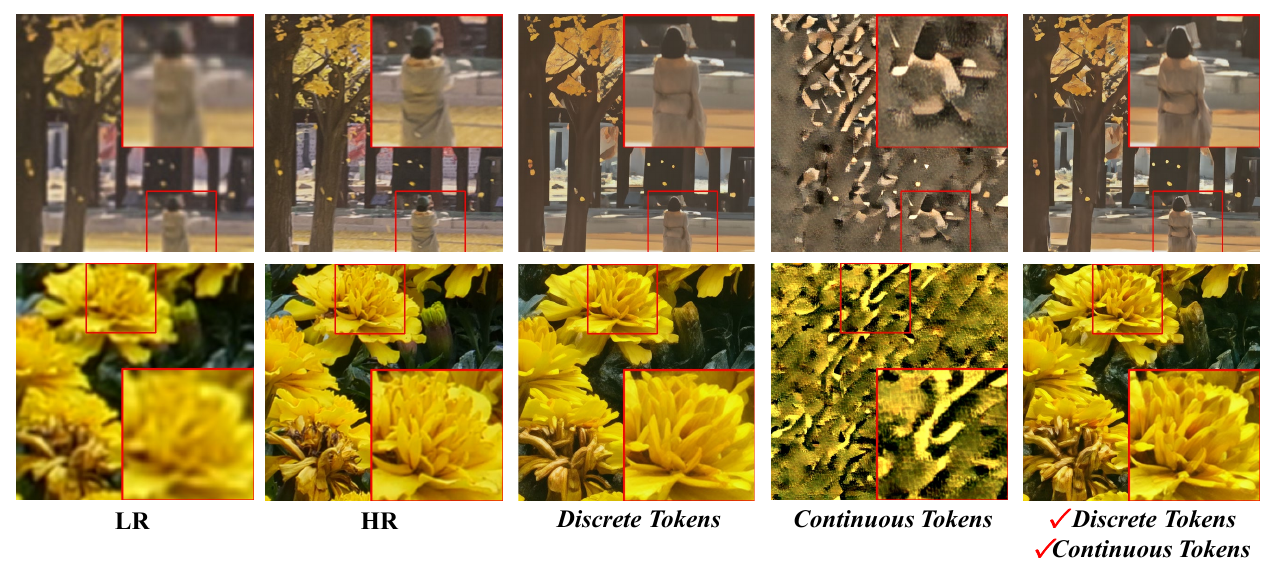}
        \vspace{-5mm}
        \caption{Visualization of \textit{Continuous Tokens} and \textit{Discrete Tokens}.}
        \label{fig:diff}
    \end{minipage}\\
    \begin{minipage}[t]{\linewidth}
        \centering
        \includegraphics[width=0.81\linewidth]{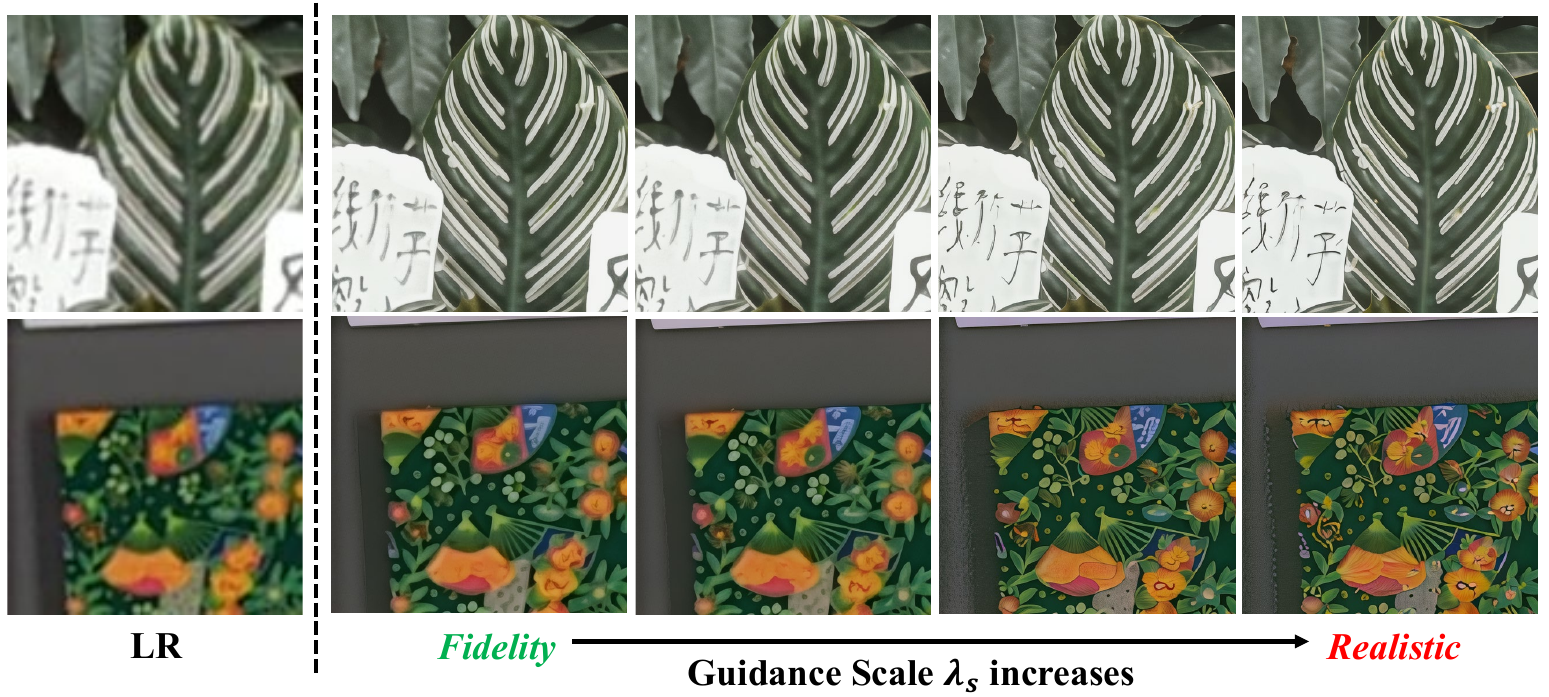}
          \vspace{-4mm}
          \caption{Effect of the \textit{Image-based CFG} with guidance scale $\lambda_s$.}
          \vspace{-5mm}
          \label{fig:cfgpic}
    \end{minipage}
\end{figure}

\subsection{Ablation Study}
\paragraph{\textbf{\textit{LR condition.}}}
We use the form of \textit{Prefix Tokens} to provide conditional control from LR images, comparing this mode with the other three LR condition modes: (1) Adding LR features directly in the transformer; (2) ControlNet \cite{zhang2023adding} with a 1/2 depth; (2) ControlNet combined with cross-attention (CA) \cite{yang2025pixel}.
In Tab. \ref{tab:lrcondition}, \textit{Prefix Tokens} mode outperforms in all metrics.
In Fig. \ref{fig:lrcondition}, our method is the only one capable of generating the correct traffic light colors and clear text.
These results showcase the effectiveness of the proposed \textit{Prefix Tokens} pattern in extracting sufficient LR information for conditional guidance.

\vspace{-1mm}
\paragraph{\textbf{\textit{Scale-aligned RoPE.}}}

We validate the effectiveness of \textit{SA-RoPE} by contrasting it with two scenarios: (1) Utilizing the original APE; (2) Applying \textit{SA-RoPE} solely on discrete tokens. 
In Tab. \ref{tab:rope}, following the gradual application of \textit{SA-RoPE}, fidelity metrics have shown a significant improvement.
In Fig. \ref{fig:rope}, it is evident that applying \textit{SA-RoPE} can better preserve the spatial structure, generating clear text and architectural details faithful to the original image.


\paragraph{\textbf{\textit{Diffusion Refiner.}}}
We use the \textit{Diffusion Refiner} to generate continuous tokens as the prediction of quantization residual loss, supplementing the discrete tokens of the autoregressive transformer.
In Tab. \ref{tab:diff}, after adding the \textit{Diffusion Refiner}, both fidelity and image quality have significantly improved.
In Fig. \ref{fig:diff}, we visualize discrete tokens and continuous tokens, showing that continuous tokens can effectively capture high-frequency details lost during quantization, resulting in images with richer semantic textures.

\paragraph{\textbf{\textit{Image-based CFG.}}}
In Sec. \ref{sec:cfg}, we introduce the \textit{Image-based CFG} to generate more realistic and higher-quality images.
Tab. \ref{tab:cfg} shows a notable enhancement in perceptual quality metrics, after incorporating the CFG.
In Fig. \ref{fig:cfgpic}, we show visualizations where images become clearer with rich textures as the guidance scale $\lambda_s$ increases. 
However, excessive $\lambda_s$ values may introduce artifacts not present in the original image. 
Therefore, our \textit{Image-based CFG} achieves a balance between fidelity and realism by controlling $\lambda_s$. 
In practice, $\lambda_s=6.0$ is optimal, and it should increase linearly with scale for finer texture details in larger scales.


\section{Conclusion}
We explored VAR in ISR tasks and proposed the VARSR framework.
To ensure pixel-level fidelity and realism, we made improvements in LR conditioning, structure representation, quantization prediction, and CFG guidance.
We collected a large-scale dataset and designed a training process.
Extensive experiments validate the performance of VARSR in generating high-fidelity and high-quality images.



\clearpage

\section*{Impact Statement}
This paper introduces the visual autoregressive into the image super-resolution field.
Therefore, any potential societal consequences or impacts related to ISR tasks apply here, as our work introduces new ideas that enhance ISR tasks with high efficiency and practicality.

\bibliography{main}
\bibliographystyle{icml2025}

\newpage
\appendix
\onecolumn
\section{Implementation Details}
\subsection{Details of VARSR framework}
\begin{table*}[h]
  \centering
  \caption{Reconstruction results of the VQVAE on \textit{RealSR} dataset.}
  \label{tab:vqvae}
  \footnotesize
  \begin{tabular}{c|c|cccccccc}
    \toprule
    Methods& Compression ratio &  PSNR $\uparrow$ & SSIM$\uparrow$ & LPIPS$\downarrow$ &DISTS$\downarrow$ & FID$\downarrow$ &  MANIQA $\uparrow$ & CLIPIQA$\uparrow$\\
    \midrule
    VQVAE (w/ quantization) & 16 & 34.33&0.9380 & 0.0428 & 0.0484 & 17.19& 0.3432& 0.4615
 \\
    VQVAE (w/o quantization) & 16 & 29.86 & 0.9006 &0.0752 &0.0688 &34.27 &0.3172 &0.4327
\\
  \bottomrule
  \end{tabular}
\end{table*}

\paragraph{\textbf{\textit{VQVAE Tokenizer.}}} 
We employ VQVAE \cite{van2017neural} as the discrete tokenizer for VARSR, following the basic settings of VAR \cite{var}.
The downsampling factor of VQVAE is $16\times$, and all scales share the same codebook with size $|V|=4096$.
Due to the diffusion refiner in our VARSR, which can predict quantization loss to some extent, we follow the alternating training strategy of Hart \cite{tang2024hart}. 
This involves incorporating continuous tokens for reconstruction with a probability of 50\% (\textit{i.e.}, bypassing the quantization process), ensuring that VQVAE can reconstruct images in both scenarios.
The reconstruction results in both scenarios are illustrated in Tab. \ref{tab:vqvae}.

\paragraph{\textbf{\textit{Image Encoder.}}}
We use a pyramid-style image encoder to extract information from degraded LR images and downsample it by a factor of 16 to obtain $32\times32$ feature space vectors.
Therefore, the image encoder consists of 4 layers, each composed of two convolutional layers, including the last one with a stride of $2\times2$ to reduce the size of the feature maps by half.

\paragraph{\textbf{\textit{Autoregressive Transformer.}}}
We utilize a Transformer based on the standard GPT-2 \cite{radford2019language} architecture with a depth of 24 layers and a width of 1536.
The modulate layer adopts the standard AdaLN \cite{peebles2023scalable} form, which is added separately in each block to generate the offsets.
The tokens are divided into 10 scales corresponding to resolutions ranging from $16\times16$ to $512\times512$, with prefix tokens from LR added at the beginning.
In the attention calculation, queries and keys reflect spatial positional information through our Scale-aligned RoPE.
During inference, KV-cache \cite{shazeer2019fast} is utilized to enhance the speed. 
Our VARSR supports the generation of higher-resolution images by concatenating them in a tiled manner.

\paragraph{\textbf{\textit{Diffusion Refiner.}}}
Our Diffusion Refiner adheres to the specifications of \cite{li2024autoregressive}, naturally supporting applications in CFG form.
The Diffusion Refiner comprises a highly lightweight residual MLP \cite{he2016deep}, with each module consisting of a LayerNorm, a linear layer, and an activation layer.
Our model consists of 6 layers, with a channel dimension of 1024. The hidden states from the autoregressive transformer serve as conditional controls, and are introduced alongside timestamps $T$ through the modulate module.
Our noise schedule follows a cosine shape, with 1000 steps during training, and is resampled with 10 steps during inference, thus requiring minimal inference time.
The parameter count of the Diffusion Refiner is less than 40M.

\subsection{Details of Large-scale dataset}

\begin{figure}[h]
  \centering
    \includegraphics[width=0.67\linewidth]{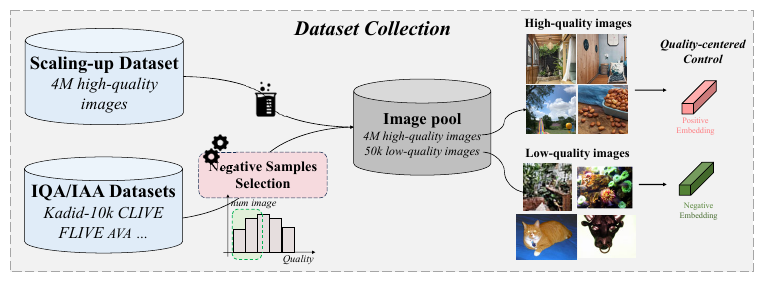}
      \vspace{-0.3cm}
  \caption{Databaet Collection. We collect a large-scale, high-quality dataset, and select 50k low-quality images from IQA and IAA datasets as negative samples, each with different \textit{quality-centered controls} attached.}
  \label{fig:data}
  \vspace{-0.3cm}
\end{figure}
\begin{table*}[h]
  \centering
  \caption{Comparison of our large-scale dataset with other well-known datasets.}
  \label{tab:dataset}
  \footnotesize
  \begin{tabular}{c|c|c|cccc}
    \toprule
    Dataset & Scale & Resolution & MANIQA $\uparrow$ &CLIPIQA$\uparrow$ & MUSIQ$\uparrow$ & Aesthetic Score$\uparrow$\\
    \midrule
    DIV2K & 800& 2k &0.43 & 0.62&71.15 & 5.57\\
    Flickr2K & 2650 & 2k & 0.41&0.58 & 70.83&5.22   \\
    \textbf{Large-scale data} & \textbf{Over 4M} & 2k & \textbf{0.48}&\textbf{0.66}&\textbf{72.31} &\textbf{5.62}\\
  \bottomrule
  \end{tabular}
\end{table*}

We collect billions of images from public datasets (e.g., LAION \cite{LAION_2022}, DataComp \cite{DataComp_2023}) and internal datasets. Employing a progressive filtering and semantic balance strategy, we have constructed a dataset comprising 4 million high-quality, high-resolution images.

\paragraph{\textbf{\textit{Progressive Filtering.}}} Images are progressively filtered using the following sub-metrics, with thresholds set for each: image metadata (including resolution, aspect ratio, and bits per pixel), IQA scores (comprising MANIQA \cite{DBLP:conf/cvpr/YangWSLGCWY22}, MUSIQ \cite{ke2021musiq}, CLIPIQA \cite{wang2023exploring}, and Aesthetic score \cite{LAIONAes_2022}), and texture richness (evaluated by the ratio of the power spectrum of high-frequency components \cite{HQ50K_2023} and the detection of blurred and flat regions \cite{LSDIR_2023}). We establish the threshold for each metric at the top 70th percentile of its distribution within the DIV2K \cite{DBLP:conf/cvpr/AgustssonT17} and Flickr2K \cite{timofte2017ntire} datasets, ensuring that the images exhibit abundant details and high-frequency textures.

\paragraph{\textbf{\textit{Semantic Balance.}}} To achieve diversity and balance of images across various domains, we implement semantic clustering based on the CLIP \cite{CLIP_2021} and SigLIP \cite{SigLIP_2023} models, and quantitatively select a sufficient number of images according to our predefined semantic categories, while specifically supplementing those categories with fewer images. Consequently, the dataset encompasses a diverse range of scenes, including portraits, people, food, animals, natural landscapes, cartoons, urban landscapes, and indoor and outdoor scenes, thereby ensuring comprehensive coverage of visual concepts and richness in scene content.

\section{Additional Experimental Results}

\subsection{Ability of the Base Model}
\begin{table*}[h]
  \centering
  \caption{Comparison of the base generative model on C2I task.}
  \label{tab:basemodel}
  \footnotesize
  \begin{tabular}{c|c|ccc}
    \toprule
    Methods& Dataset &  MANIQA $\uparrow$ &CLIPIQA$\uparrow$ & MUSIQ$\uparrow$ \\
    \midrule
    Original VAR & ImageNet & 0.3250 &0.5496 &60.92\\
    Original VAR & Large-scale data &0.4852&0.7015&72.74\\
    VARSR & Large-scale data & \textbf{0.5634} &  \textbf{0.7314} &  \textbf{74.32}\\
  \bottomrule
  \end{tabular}
\end{table*}
\begin{figure*}[h]
  \centering
    \includegraphics[width=0.9\linewidth]{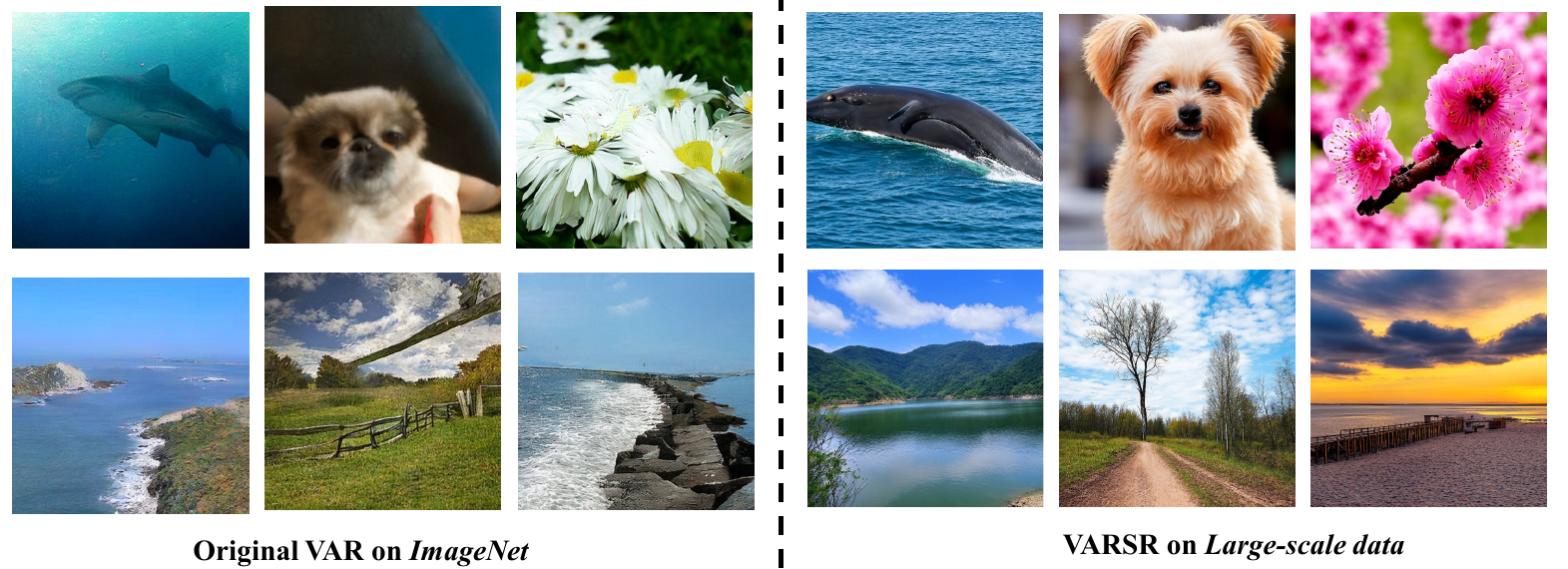}
  \caption{Visualization of C2I generation results from closely related categories.}
  \label{fig:basemodel}
\end{figure*}
To enhance the generative capabilities of the base model in the C2I task for downstream task migration, we pre-train the VAR model on our large-scale high-quality dataset.
In this section, we compare the capabilities of our VARSR base model trained on our large-scale dataset with the original VAR model \cite{var} trained on ImageNet \cite{deng2009imagenet}.
As we are particularly interested in the model's ability to generate detailed textures and ensure that it can still produce richly detailed images for downstream ISR task migration, we use non-reference IQA metrics for evaluation.
In addition, it is unfair to use reference metrics such as FID for testing due to the differences in datasets.

Quantitative results are shown in Tab. \ref{tab:basemodel}, and in Fig .\ref{fig:basemodel}, we visualize some generated results from closely related categories.
After using our high-quality large-scale dataset, the generated images show significant improvements in both objective metrics and subjective human observations.
Furthermore, by incorporating our improvements (\textit{e.g.}, Scale-aligned RoPE, Diffusion Refiner) into the original VAR model, \textit{i.e.}, using the VARSR model, the quality further improved. 
This validates the effectiveness of our enhancements for the original C2I task.
The results above validate the capabilities of our base model, which can generate images with rich details and clear semantics. It possesses a strong generative prior that can be leveraged by ISR tasks.

\subsection{Comparisons with SOTA Methods}
\begin{table}[h]
  \centering
  \footnotesize
  \setlength{\tabcolsep}{5pt}
  \caption{Results of user study on real-world images.}
  \label{tab:userstudy}
  \centering
  \begin{tabular}{c|cccccc}
    \toprule
        Methods & BSRGAN & Real-ESRGAN  & StableSR & PASD & SeeSR & \textbf{VARSR(Ours)} \\
    \midrule
    Selection Rates & 0.5\% & 0.9\% &  4.6\% & 18.5\% & 20.3\% & \textbf{55.2\%}\\
  \bottomrule
  \end{tabular}
\end{table}

\paragraph{\textbf{\textit{User Study.}}} 
In order to comprehensively assess the performance of our VARSR in real-world scenarios, we conduct a user study on 50 randomly sampled LR real-world images from \textit{DrealSR}\cite{DBLP:conf/iccv/CaiZYC019} and \textit{RealSR}\cite{DBLP:conf/eccv/WeiXLZYZL20}.
We compare our VARSR with five other GAN-based and diffusion-based ISR methods, including BSRGAN~\cite{DBLP:conf/iccv/0008LGT21}, Real-ESRGAN~\cite{DBLP:conf/iccvw/WangXDS21}, StableSR~\cite{DBLP:journals/corr/abs-2305-07015}, PASD~\cite{yang2025pixel} and SeeSR \cite{DBLP:journals/corr/abs-2311-16518}.
For each image, participants were presented with both the LR image and the restoration results of all ISR methods, and were then asked to indicate their choice for the best ISR result for the LR image.
We invited 20 visual researchers to participate in the user study, and in total, we obtained $20\times 50$ selection results.
As shown in Tab. \ref{tab:userstudy}, our VARSR achieves \textbf{the highest selection rate of 55.2\%}, far surpassing other methods, demonstrating the powerful capability of VARSR in real-world scenarios to generate realistic images that align with human aesthetics.

\begin{table}[h]
  \centering
  \footnotesize
  \caption{Quantitative comparison with SOTA methods on \textit{RealLR200} dataset with no reference images.
  \textcolor{red}{Red} and \textcolor{blue}{blue} colors are the best and second-best performance.
  }
  \label{tab:reallr}
  \centering
  \begin{tabular}{c|c|ccc|ccccc|c}
    \toprule
    \multirow{2}{*}{Dataset} & \multirow{2}{*}{Metrics} & \multicolumn{3}{c|}{GAN-based} & \multicolumn{5}{c|}{Diffusion-based} & AR-based\\
    & & BSRGAN & Real-ESR & SwinIR & LDM & StableSR & DiffBIR & PASD & SeeSR & \textbf{VARSR} \\
    \midrule
    \multirow{3}{*}{\textit{RealLR200}}
    & MANIQA$\uparrow$   & 0.3671& 0.3633& 0.3741& 0.3049& 0.3688& 0.4288& 0.4295& \textcolor{blue}{0.4844}& \textcolor{red}{0.5177} \\
    & CLIPIQA$\uparrow$  & 0.5698& 0.5409& 0.5596&  0.5253&0.5935&0.6452 & 0.6325&\textcolor{blue}{0.6553} & \textcolor{red}{0.7513} \\
    & MUSIQ$\uparrow$    & 64.87& 62.96& 63.55& 55.19&  63.29 & 62.44 & 66.50& \textcolor{blue}{ 68.37} & \textcolor{red}{71.92} \\
  \bottomrule
  \end{tabular}
\end{table}
\paragraph{\textbf{\textit{Comparisons on real-world images}}} 
To evaluate the performance of our VARSR in in-the-wild scenarios, we test different approaches on the \textit{RealLR200} dataset \cite{DBLP:journals/corr/abs-2311-16518}, which comprises 200 real-world images collected from previous studies \cite{DBLP:journals/corr/abs-2308-15070, DBLP:conf/iccvw/WangXDS21} and from the internet.
Due to the absence of available reference HR images, we utilize only three no-reference IQA metrics: MANIQA \cite{DBLP:conf/cvpr/YangWSLGCWY22}, MUSIQ \cite{ke2021musiq}, and CLIPIQA \cite{wang2023exploring}.

\begin{figure*}[h]
  \centering
    \includegraphics[width=0.88\linewidth]{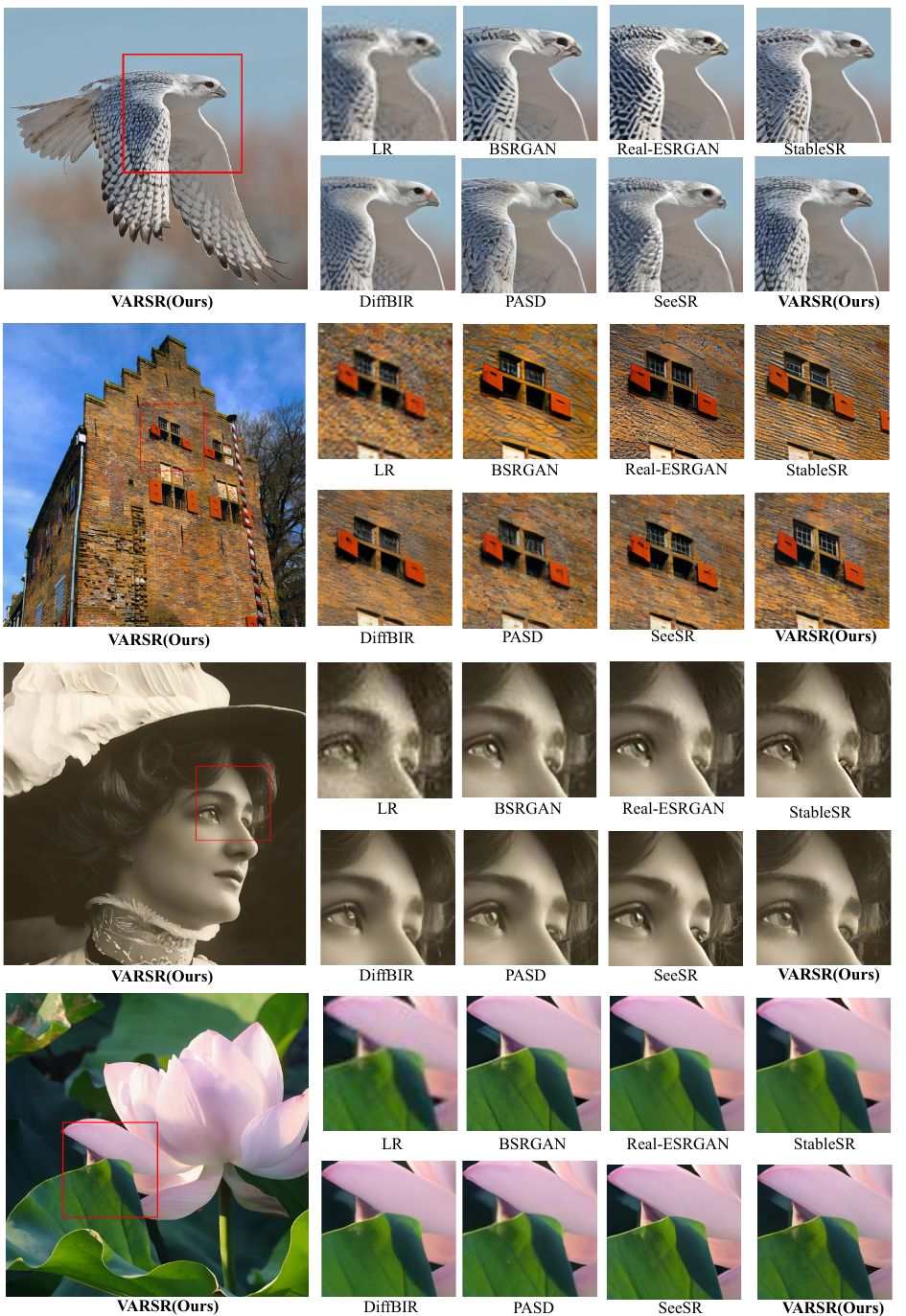}
  \caption{Comparisons with different methods on real-world images. \textbf{Zoom in for a better view}.}
  \label{fig:realimage}
\end{figure*}

As shown in Tab. \ref{tab:reallr}, our VARSR achieves  \textbf{the best performance among all metrics}, which is consistent with the results on other synthetic and real-world datasets.
In Fig. \ref{fig:realimage}, we visualize some ISR results, demonstrating that VARSR can generate images with more realistic details compared to other methods.
The results above validate the strong restoration capability of VARSR in real-world scenarios, showcasing its practical application value.

\subsection{Ablation Study}
\paragraph{\textbf{\textit{Large-scale Dataset.}}}
\begin{table}[h]
	\begin{minipage}[t]{0.48\linewidth}
		\centering
        \footnotesize
        \caption{Ablation on the \textit{Training Database}.}
		  \label{tab:data}
   \centering
   \resizebox{1.0\linewidth}{!}{
  \begin{tabular}{c|cc|cc}
    \toprule
    \multirow{2}{*}{Metrics} &\multicolumn{2}{c|}{\textit{DrealSR}} & \multicolumn{2}{c}{\textit{RealSR}} \\
    & DIV2K & Scaling-up & DIV2K &  Scaling-up\\
    \midrule
    SSIM$\uparrow$ & \textbf{0.7750} & 0.7652 & \textbf{0.7284} & 0.7169\\
    LPIPS$\downarrow$ & \textbf{0.3441} &0.3541 & \textbf{0.3376} &0.3504 \\
    DISTS$\downarrow$ & \textbf{0.2434} & 0.2526 & \textbf{0.2321} &0.2470 \\
    MANIQA$\uparrow$ & 0.4885 & \textbf{0.5362}  & 0.4990 &\textbf{0.5570} \\
    MUSIQ$\uparrow$ & 64.68 &\textbf{68.15} & 67.36  & \textbf{71.26} \\
  \bottomrule
  \end{tabular}}
	\end{minipage}
	\hfill 
	\begin{minipage}[t]{0.48\linewidth}
		\centering
		\footnotesize
  \caption{Ablation on the \textit{Image Preprocess Strategy}.}
  \label{tab:preprocess}
   \resizebox{1.0\linewidth}{!}{
  \begin{tabular}{c|cc|cc}
    \toprule
    \multirow{2}{*}{Metrics} &\multicolumn{2}{c|}{\textit{DrealSR}} & \multicolumn{2}{c}{\textit{RealSR}} \\
    & Rcrop & Resize\&Crop & Rcrop & Resize\&Crop \\
    \midrule
    SSIM$\uparrow$ & \textbf{0.7772} & 0.7652 & \textbf{0.7304} & 0.7169\\
    LPIPS$\downarrow$ & \textbf{0.3279} &0.3541 & \textbf{0.3317} &0.3504 \\
    DISTS$\downarrow$ & \textbf{0.2335} & 0.2526 & \textbf{0.2286} &0.2470 \\
    MANIQA$\uparrow$ & 0.4577 & \textbf{0.5362}  & 0.4639 &\textbf{0.5570} \\
    MUSIQ$\uparrow$ & 63.19 &\textbf{68.15} & 66.62  & \textbf{71.26} \\
  \bottomrule
  \end{tabular}}
	\end{minipage}
\end{table}
\begin{figure}[h]
  \centering
    \includegraphics[width=0.6\linewidth]{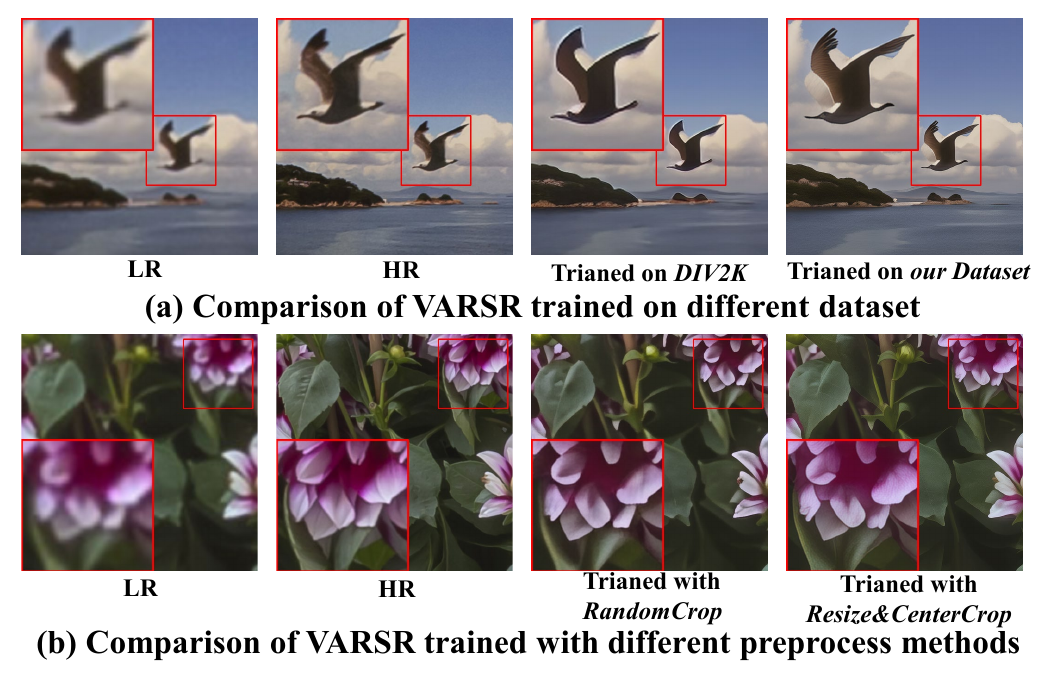}
    \vspace{-3mm}
  \caption{Comparison of \textit{different dataset} and \textit{image preprocess methods}.}
    \vspace{-3mm}
  \label{fig:dataexp}
\end{figure}
In Tab. \ref{tab:data}, we validate the importance of training VARSR with large-scale high-quality data.
We separately test the model trained on our large-scale dataset (over 4M images) and the model trained using the DIV2K \cite{DBLP:conf/cvpr/AgustssonT17} series dataset (12k images, following the setting of \cite{yang2025pixel}).
Training with a large amount of high-quality data results in a significant performance improvement, as also confirmed by the visual results in Fig. \ref{fig:dataexp} (a). 
This demonstrates the effectiveness and necessity of using large-scale high-quality data.

\paragraph{\textbf{\textit{Image Preprocess.}}}
As depicted in Sec. \ref{sec:data}, we preprocess the training images by first resizing and then cropping to include more foreground semantic information.
In Tab. \ref{tab:preprocess} and Fig. \ref{fig:dataexp} (b), we compare this preprocessing method with the commonly used random cropping strategy.
Our strategy achieves higher perceived quality, producing richer details, which demonstrates that high-frequency semantic information can be better preserved through our preprocessing.

\paragraph{\textbf{\textit{VQVAE Dropout Strategy.}}}
\begin{table*}[h]
  \footnotesize
  \centering
  \caption{Ablation on the \textit{VQVAE Scale Dropout Strategy}.}
  \label{tab:vae}
   \centering
  \begin{tabular}{c|cc|cc}
    \toprule
    \multirow{2}{*}{Metrics} &\multicolumn{2}{c|}{\textit{DrealSR}} & \multicolumn{2}{c}{\textit{RealSR}} \\
    & w/o & $p_d=0.1$ & w/o & $p_d=0.1$ \\
    \midrule
    SSIM$\uparrow$ & 0.7567 & \textbf{0.7652} & 0.7155 & \textbf{0.7169}\\
    LPIPS$\downarrow$ & 0.3667 &\textbf{0.3541} & 0.3587 &\textbf{0.3504} \\
    DISTS$\downarrow$ & 0.2634 & \textbf{0.2526} & 0.2511&\textbf{0.2470} \\
    MANIQA$\uparrow$ & \textbf{0.5410} & 0.5362  & 0.5414 &\textbf{0.5570} \\
    MUSIQ$\uparrow$ & 67.98 &\textbf{68.15} & 70.83  & \textbf{71.26} \\
  \bottomrule
  \end{tabular}
\end{table*}
\begin{figure*}[h]
  \centering
    \includegraphics[width=0.6\linewidth]{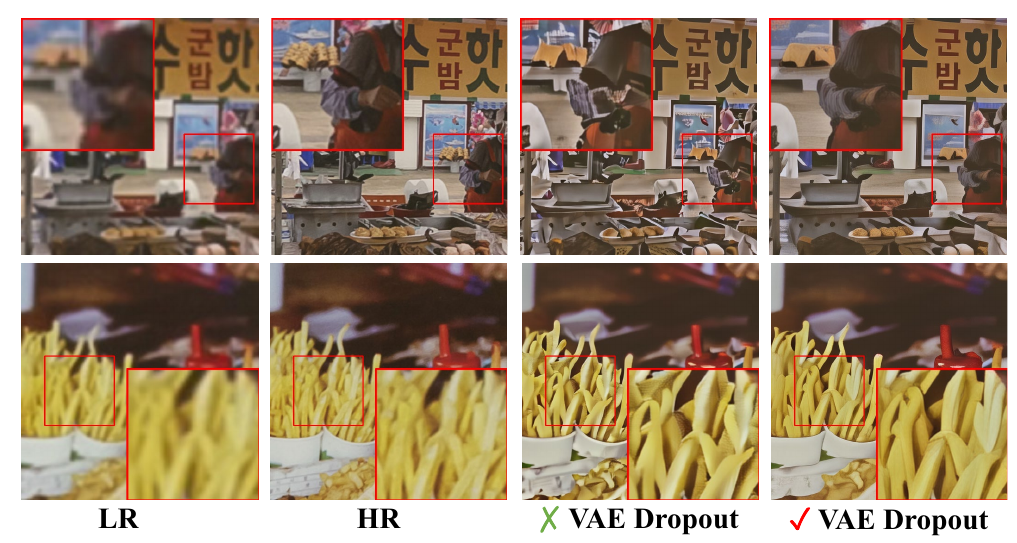}
  \vspace{-3mm}
  \caption{Effectivess of the \textit{VQVAE Scale Dropout Strategy}.}
  \vspace{-3mm}
  \label{fig:vae}
\end{figure*}
In Tab. \ref{tab:vae}, we compare the results of training the VARSR with the VQVAE tokenizer under two conditions, representing whether the VQVAE uses the \textit{scale dropout strategy} during training with a dropout ratio $p_d=0.1$.
The tokenizer using the dropout strategy achieves significant improvements in all metrics, particularly in fidelity-based metrics, as the preceding scales can provide richer semantic information.
As shown in Fig. \ref{fig:vae}, when the scale dropout strategy is not applied, all the fine texture details are generated in the final scale,  which may result in some undesired artifacts in generated images.
The condition with the dropout strategy applied can consistently generate images that are faithful to the original image.

\subsection{Limitations}
\begin{figure*}[h]
  \centering
    \includegraphics[width=0.9\linewidth]{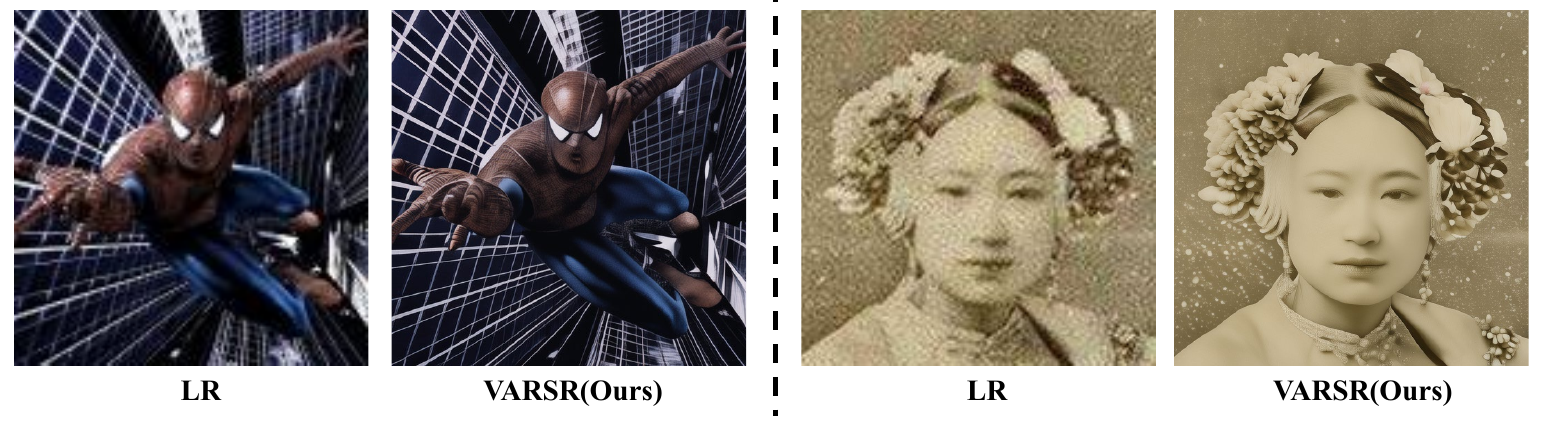}
  \caption{Limitations of VARSR. Due to limitations in the scope of training data coverage, it may be challenging to restore the correct semantics in certain extreme scenarios.}
  \label{fig:limitations}
\end{figure*}


Although our VARSR base model is pretrained on the large-scale dataset of 4M images, the semantic coverage may still have some limitations compared to the billions of data used by Stable Diffusion  \cite{rombach2022high}. 
When the original image contains rare semantics and is severely damaged, it may not be possible to correctly restore the content.
As shown in Fig. \ref{fig:limitations}, the LR image's severe distortion impedes the accurate restoration of the famous character Spider-Man in the left case, while the generated image of an ancient woman in the right case features simplistic and unrealistic head adornments.

\section{More Visualization Results}
\begin{figure}[t]
  \centering
    \includegraphics[width=0.88\linewidth]{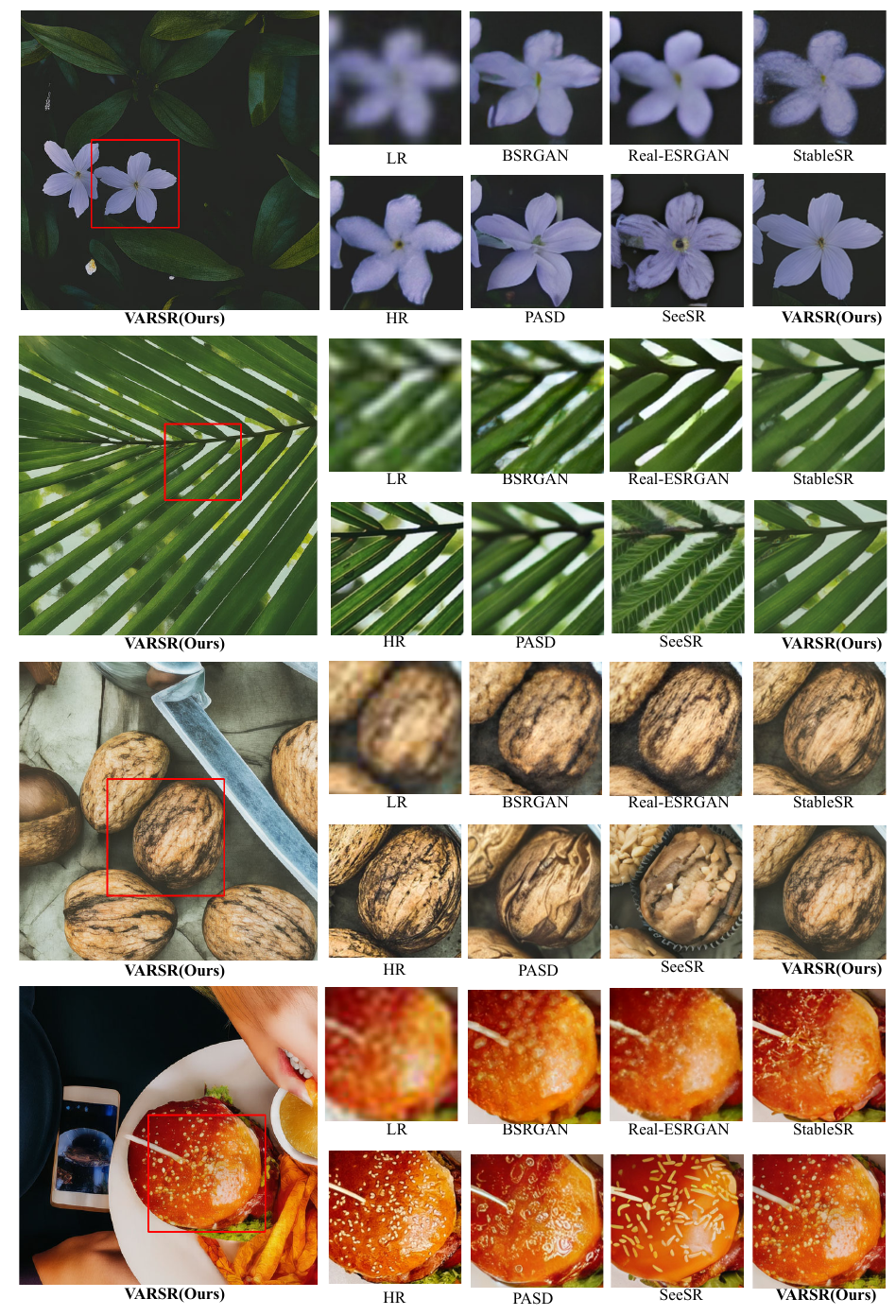}
  \caption{Additional qualitative comparisons with different SOTA methods (Part 1). \textbf{Zoom in for a better view}.}
  \label{fig:sota_add0}
\end{figure}

\begin{figure}[t]
  \centering
    \includegraphics[width=0.88\linewidth]{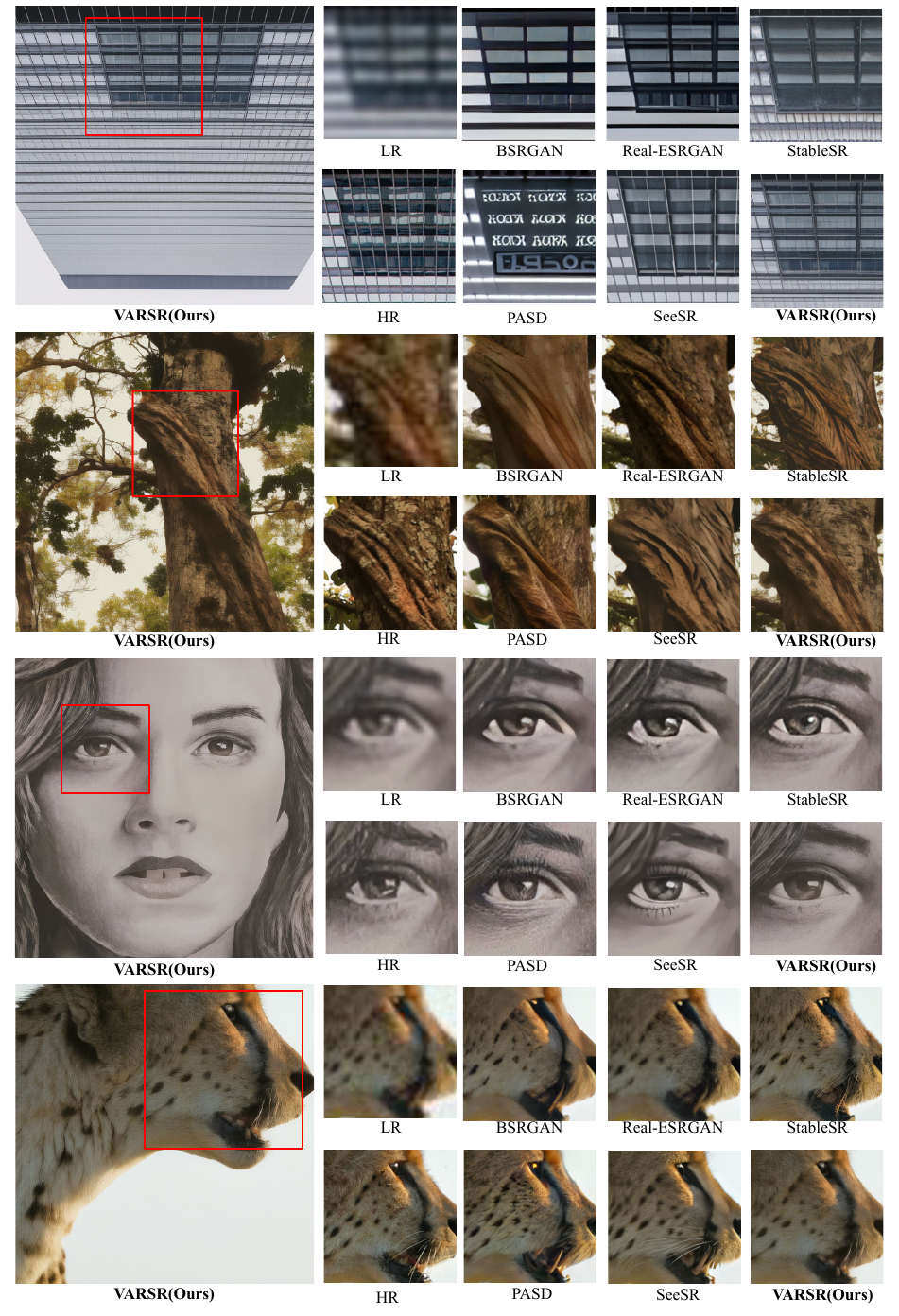}
  \caption{Additional qualitative comparisons with different SOTA methods (Part 2). \textbf{Zoom in for a better view}.}
  \label{fig:sota_add1}
\end{figure}

In Fig. \ref{fig:sota_add0} and Fig. \ref{fig:sota_add1}, we provide additional comparison visualization results with other methods, showcasing the robust capability of VARSR in generating high-fidelity and high-realistic images.
\clearpage


\end{document}